%% file: iclr19_coex.tex
\pdfoutput=1
\documentclass{article} %
\usepackage{iclr,times}

\iclrfinalcopy %

  \usepackage{lineno}

  \usepackage{marginnote}
  
\usepackage[utf8]{inputenc} %
\usepackage[T1]{fontenc}    %
\usepackage{url}            %
\usepackage{booktabs}       %
\usepackage{amsfonts}       %
\usepackage{nicefrac}       %
\usepackage{microtype}      %

\usepackage{times}
\usepackage{epsfig}
\usepackage{graphicx}
\usepackage{cleveref}
\usepackage{amsmath}
\usepackage{amssymb}
\usepackage{bbm}
\usepackage{lipsum}
\usepackage{xspace}

\usepackage{array}
\usepackage{comment}
\usepackage{kotex}
\usepackage{enumerate}
\usepackage{enumitem}
\usepackage{adjustbox}
\usepackage{caption}
\usepackage{subfigure}
\usepackage{wrapfig}

\usepackage{morefloats}
\usepackage{dblfloatfix}
\usepackage{placeins}
\usepackage{makecell}

\usepackage{algorithm}
\usepackage{algorithmic}
\usepackage{color}
\PassOptionsToPackage{usenames,dvipsnames}{xcolor}

\newcommand{\alert}[1]{{\color{red}{#1}}}

\newcommand{\argmax}{\operatornamewithlimits{argmax}}

\newcommand{\softmax}{\operatornamewithlimits{softmax}}

\usepackage[colorinlistoftodos]{todonotes}

\newcommand{\coex}{{CoEX}}
\newcommand{\coexRAM}{{CoEX+RAM}}
\newcommand{\ADM}{{ADM}}

\DeclareRobustCommand\onedot{\futurelet\@let@token\@onedot}
\def\onedot{.}
\def\eg{\emph{e.g}\onedot} 
\def\ie{\emph{i.e}\onedot}

\newcommand{\R}{\mathbb R}

\makeatletter
\newcommand*{\centerfloat}{%
  \parindent \z@
  \leftskip \z@ \@plus 1fil \@minus \textwidth
  \rightskip\leftskip
  \parfillskip \z@skip}
\makeatother

\PassOptionsToPackage{sort}{natbib}
\usepackage{natbib}
\setcitestyle{authoryear,round,citesep={;},aysep={,},yysep={;}}
\usepackage[pagebackref=false,breaklinks=true,colorlinks,bookmarks=false,citecolor=blue]{hyperref}

\title{
Contingency-Aware Exploration in \\ Reinforcement Learning
}

\newcommand{\ProjectURL}{https://coex-rl.github.io/}

\author{%
    Jongwook Choi\thanks{Equal contributions, listed in alphabetical order.}\,\,$^{,1}$ \hspace{1em}
    Yijie Guo$^{*,1}$          \hspace{1em}
    Marcin Moczulski$^{*,2}$          \hspace{1em} \\ \bf
    Junhyuk Oh$^{1,}\thanks{Now at DeepMind.}$          \hspace{1em}
    Neal Wu$^2$          \hspace{1em}
    Mohammad Norouzi$^2$          \hspace{1em}
    Honglak Lee$^{2,1}$      \hspace{1em}
    \\[2pt]
    $^1$University of Michigan \hspace{2em}
    $^2$Google Brain \hspace{2em}
    \\
    \texttt{{\rm\{}%
        jwook,guoyijie%
    {\rm\}}@umich.edu}
    ~~~
    \texttt{%
        moczulski%
    @google.com}\\
    \texttt{{\rm\{}%
        junhyuk,nealwu,mnorouzi,honglak%
    {\rm\}}@google.com} \\
}



\begin{document}

\maketitle

\newcommand{\MontezumaRevenge}{\textsc{Montezuma's Revenge}\xspace}
\newcommand{\Gravitar}{\textsc{Gravitar}\xspace}
\newcommand{\Seaquest}{\textsc{Seaquest}\xspace}
\newcommand{\Venture}{\textsc{Venture}\xspace}
\newcommand{\PrivateEye}{\textsc{PrivateEye}\xspace}
\newcommand{\Hero}{\textsc{Hero}\xspace}
\newcommand{\Freeway}{\textsc{Freeway}\xspace}
\newcommand{\Frostbite}{\textsc{Frostbite}\xspace}
\newcommand{\Qbert}{\textsc{Qbert}\xspace}

\vspace*{-.3cm}
\begin{abstract}
\vspace*{-.2cm}

\input{00_abstract}
\end{abstract}

\input{01_introduction}

\input{02_relatedwork}
\input{03_method}

\input{04_experiments}

\input{05_conclusion}
\bibliography{references}
\bibliographystyle{iclr}

\clearpage

\input{10_appendix}

\input{11_additional}
\end{document}

%% file: 00_abstract.tex
\vspace*{-2pt}
This paper investigates whether learning contingency-awareness and controllable aspects of an environment can lead to better exploration in reinforcement learning.
To investigate this question, we consider an instantiation of this hypothesis evaluated on the Arcade Learning Element (ALE). %
In this study, we develop an attentive dynamics model (\ADM{}) that discovers controllable elements of the observations, which are often associated with the location of the character in Atari games.
The \ADM{} is trained in a self-supervised fashion 
to predict the actions taken by the agent. %
The learned contingency information is used as a part of the state representation for exploration purposes.
We demonstrate that combining actor-critic algorithm with count-based exploration using %
our representation achieves impressive results on
a set of notoriously challenging Atari games due to sparse rewards.\footnote{%
Examples of the learned policy and the %
contingent regions are available at
\url{\ProjectURL}.
}
For example, we report a state-of-the-art score of
>11,000
points on \MontezumaRevenge without using expert demonstrations, explicit high-level information (\eg, RAM states), or supervisory data.
Our experiments confirm that contingency-awareness is indeed an extremely powerful concept for tackling exploration problems in reinforcement learning
and opens up interesting research questions for further investigations.

%% file: 01_introduction.tex
\section{Introduction}
\label{sec:introduction}
\vspace*{-0.15cm}

The success of reinforcement learning (RL) algorithms in complex environments hinges on the way
they balance {\em exploration} and {\em exploitation}.
There has been a surge of recent interest in developing effective exploration strategies for problems
with high-dimensional state spaces and sparse rewards %
\citep{Schmidhuber:1991,Oudeyer:2009,Houthooft:NIPS2016:VIME,Bellemare:NIPS2016:UnifyingCount,Osband:NIPS2016:BootstrappedDQN,Pathak:ICML2017:Curiosity,Plappert:ICLR2018:ParamExp,Zheng:NIPS2018:LIRPG}.
Deep neural networks have seen great success as expressive function approximators within RL and as
powerful representation learning methods for many domains. 
In addition, there have been recent studies on using neural network representations for exploration~\citep{Tang:NIPS2017:SimHash,Martin:1706.08090,Pathak:ICML2017:Curiosity}.
For example, count-based exploration with neural density estimation~\citep{Bellemare:NIPS2016:UnifyingCount,Tang:NIPS2017:SimHash,Ostrovski:ICML2017:ExplorationDensity} presents one of the state-of-the-art techniques on the most challenging Atari games with sparse rewards. 

Despite the success of recent exploration methods, it is still an open question on how to construct an optimal representation for exploration. 
For example, the concept of visual similarity is used for learning density models as a basis for calculating \emph{pseudo-counts}~\citep{Bellemare:NIPS2016:UnifyingCount,Ostrovski:ICML2017:ExplorationDensity}. 
However, as \cite{Tang:NIPS2017:SimHash} noted, 
the ideal way to represent states should be based on what is relevant to solving the MDP, rather than only relying on visual similarity.
In addition, there remains another question on whether the representations used for recent exploration works are easily interpretable. %
To address these questions, we investigate whether we can learn a complementary, more intuitive, and interpretable
high-level abstraction that can be very effective in exploration
by using the ideas of contingency awareness %
and controllable dynamics.

The key idea that we focus on in this work is the notion of \emph{contingency awareness}~\citep{watson1966:contingency,Bellemare:AAAI2012:Contingency}
--- the agent's understanding of the environmental dynamics and recognizing that some aspects of the dynamics are under the agent's control. 
Intuitively speaking, this can represent the segmentation mask of the agent
operating in the 2D or 3D environments (yet one can think of more abstract and general state spaces). 
In this study, we investigate the concept of contingency awareness based on \emph{self-localization}, \ie, the awareness of where the agent is 
located in the abstract state space.
We are interested in discovering parts of the world that are directly dependent on the agent's immediate action, which often reveal the agent's
approximate location.
For further motivation on the problem, we note that contingency awareness is a very important concept in neuroscience and psychology. %
In other words, being self-aware of one's location is an important property within many observed intelligent organisms and systems.
For example, recent breakthroughs in neuroscience, such as the Nobel Prize winning work on the grid cells~\citep{Moser:2015:GridCell,Banino:Nature2018:NavigateAI}, show that organisms that perform very well in spatially-challenging tasks are self-aware of their location. 
This allows rats to navigate, remember paths to previously visited places and important sub-goals, and find shortcuts. 
In addition, the notion of contingency awareness has been shown as an important factor in developmental psychology~\citep{watson1966:contingency,baeyens1990contingency}. 
We can think of self-localization (and more broadly self-awareness) as a principled and fundamental direction towards intelligent agents.

Based on these discussions, we hypothesize that contingency awareness can be a powerful mechanism for tackling exploration problems in reinforcement learning.
We consider an instantiation of this hypothesis evaluated on the Arcade Learning Element (ALE). 
For example, in the context of 2D Atari games, contingency-awareness roughly corresponds to understanding the notion of controllable entities (\eg, the player's avatar), which \cite{Bellemare:AAAI2012:Contingency} refer to as \emph{contingent regions}. %
More concretely, as shown in Figure \ref{fig:figure1}, in the game \Freeway, only the chicken sprite is under the agent's control and not the multiple moving cars;
therefore the chicken's location should be an informative element for exploration~\citep{Bellemare:AAAI2012:Contingency,Pathak:ICML2017:Curiosity}.
In this study, we also investigate whether contingency awareness can be learned without any external annotations or supervision. 
For this, we provide an instantiation of an algorithm for automatically learning such information %
and using it for improving exploration on a 2D ALE environment \citep{Bellemare:JAIR2013:ALE}.
Concretely, we employ an \emph{attentive dynamics model} (ADM) %
to predict the agent's action chosen between consecutive states. 
It allows us to approximate the agent's position in 2D environments, but unlike other approaches such as \citep{Bellemare:AAAI2012:Contingency}, it does not require any additional supervision to do so.
The ADM learns in an online and self-supervised fashion with pure observations as the agent's policy is updated
and does not require hand-crafted features, an environment simulator, or supervision labels for training.
In experimental evaluation, our methods significantly improve the performance of A2C on hard-exploration Atari games
in comparison with competitive methods such as density-based exploration
\citep{Bellemare:NIPS2016:UnifyingCount,Ostrovski:ICML2017:ExplorationDensity}
and SimHash \citep{Tang:NIPS2017:SimHash}.
We report very strong results on sparse-reward Atari games, including the state-of-the-art performance on the notoriously difficult  \MontezumaRevenge,
when combining our proposed exploration strategy with PPO~\citep{schulman2017:proximal},
without using expert demonstrations, explicit high-level information (\eg, RAM states),
or resetting the environment to an arbitrary state.

We summarize our contributions as follows:
\vspace*{-2pt}
\begin{itemize}[leftmargin=15pt]
    \setlength{\itemsep}{0pt}\setlength{\parskip}{2pt}
	\item We demonstrate the importance of learning contingency awareness for efficient exploration in challenging sparse-reward RL problems.
	\item We develop a novel instance of attentive dynamics model using contingency %
	and controllable dynamics %
	to provide robust localization abilities across the most challenging Atari environments. 
	\item We achieve a strong performance on difficult sparse-reward Atari games, including the state-of-the-art score %
	on the notoriously challenging \MontezumaRevenge.
\end{itemize}
Overall, we believe that our experiments confirm the hypothesis that contingency awareness is an extremely powerful concept for tackling exploration problems in reinforcement learning, which opens up interesting research questions for further investigations.

%% file: 02_relatedwork.tex
\clearpage

\section{Related Work}
\label{sec:related-work}

\vspace*{-7pt}
\paragraph{Self-Localization.}
The discovery of grid cells \citep{Moser:2015:GridCell} %
motivates working on agents that are self-aware of their location.
\cite{Banino:Nature2018:NavigateAI}
emphasize the importance of self-localization and train a neural network which learns a similar mechanism to grid cells
to perform %
tasks related to spatial navigation. The presence of grid cells is correlated with high performance.
Although grid cells seem tailored to 2D or 3D problems that animals encounter in their life,
it is speculated that their use can be extended to more abstract spaces.
A set of potential approaches to self-localization ranges from ideas specific to a given environment, \eg, SLAM \citep{Durrant:2006:SLAM}, to methods %
with potential generalizability
\citep{Mirowski:ICLR2017:Navigate,Jaderberg:ICLR2017:RLAux,Mirowski:NIPS2018:StreetLearn}.

\vspace*{-10pt}
\paragraph{Self-supervised Dynamics Model and Controllable Dynamics.}

Several works have used forward and/or inverse dynamics models of the environment \citep{Oh:NIPS2015:ActionVideoPred,Agrawal:2016,Shelhamer:2017}. %
\cite{Pathak:ICML2017:Curiosity}
employ a similar dynamics model to learn feature representations of states that captures controllable aspects of the environment.
This dense representation is used to design a curiosity-driven intrinsic reward.
The idea of learning representations on relevant aspects of the environment
by learning auxiliary tasks is also explored
in \citep{Jaderberg:ICLR2017:RLAux,Bengio:2017:FeatureControl,Sawada:2018:Controllable}.
Our presented approach is different as we focus on explicitly discovering controllable aspects using an attention mechanism, resulting in better interpretability.

\vspace*{-10pt}
\paragraph{Exploration and Intrinsic Motivation.}

The idea of providing an exploration bonus reward depending on the state-action visit-count was proposed by \cite{Strehl:2008} (MBIE-EB),
originally under a tabular setting.
Later it has been combined with different techniques %
to deal with high-dimensional state spaces. %
\cite{Bellemare:NIPS2016:UnifyingCount} use a Context-Tree Switching (CTS) density model
to derive a state \emph{pseudo-count},
whereas
\cite{Ostrovski:ICML2017:ExplorationDensity} use %
PixelCNN as a state density estimator.
\cite{Martin:1706.08090} also construct a visitation density model %
over a compressed feature space
rather than the raw observation space.
Alternatively,
\cite{Tang:NIPS2017:SimHash} propose a locality-sensitive hashing (LSH) method to cluster states
and maintain a state-visitation counter based on a form of similarity between frames.
We train an agent with a similar count-based exploration bonus,
but the way of maintaining state counter seems relatively simpler
in that key feature information (\ie, controllable region) is explicitly extracted from the observation and directly used for counting states.

\vspace*{-3pt}
Another popular family of exploration strategies in RL uses intrinsic motivation
\citep{Schmidhuber:1991,Chentanez:NIPS2004:Intrinsic,Oudeyer:2009,Barto:2013:intrinsic}. %
These methods encourage the agent to look for something surprising in the environment which motivates its search for novel states,
such as surprise~\citep{Achiam:1703.01732},
curiosity~\citep{Pathak:ICML2017:Curiosity,Burda:2018:Curiosity},
and diversity~\citep{Eysenbach:2018uc},
or %
via feature control \citep{Jaderberg:ICLR2017:RLAux,Dilokthanakul:2017:IntrinsicMotivation}.

%% file: 03_method.tex
\vspace*{-.2cm}
\section{Approach}
\vspace*{-.10cm}
\label{sec:approach}

\newcommand{\minus}{\scalebox{0.75}[1.0]{$-$}}

\vspace*{-5pt}

\setlength{\abovedisplayskip}{4pt}
\setlength{\belowdisplayskip}{3pt}
\subsection{Discovering Contingency via Attentive Dynamics Model}
\label{sec:dynamics}
\vspace*{-5pt}

\begin{figure}[tb]
\vspace*{-5pt}
\begin{center}
    \hspace*{1pt}
    \includegraphics[width=0.30\linewidth]{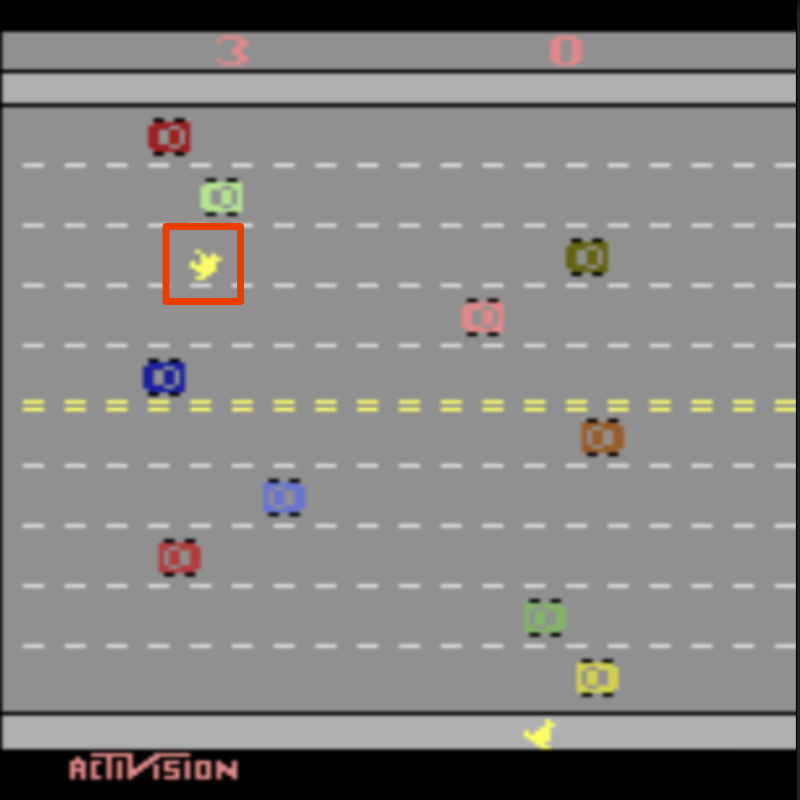}
    \hfill
    \includegraphics[width=0.65\linewidth]{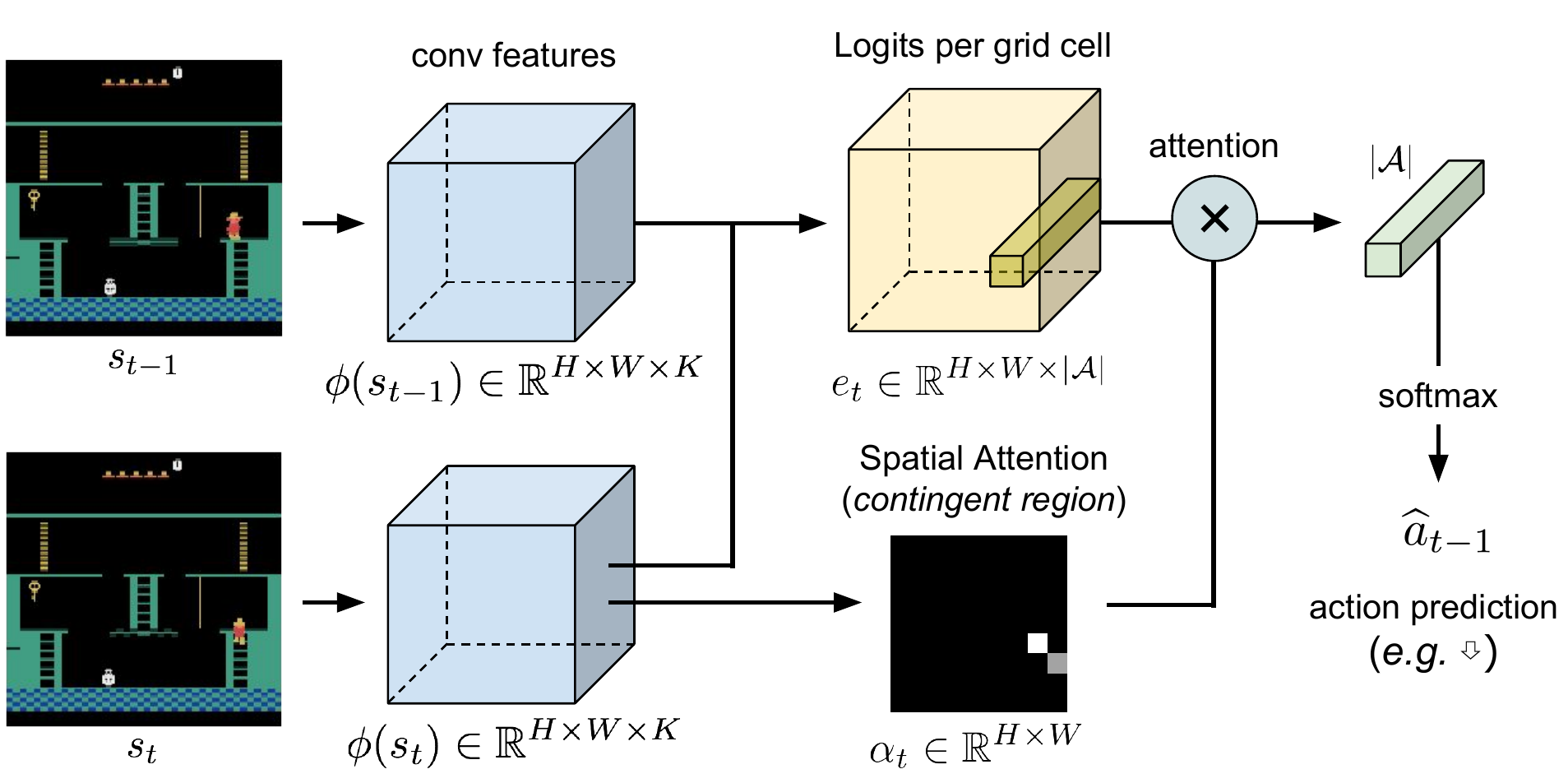}
    \caption{
        {\bf Left:} Contingent region in \Freeway; an object in a red box denotes what is under the agent's control, whereas the rest is not.
        {\bf Right:} A diagram for the proposed \ADM{} architecture.
        }
    \label{fig:figure1}
\end{center}
\vspace*{-0.4cm}
\end{figure}

To discover the region of the observation
that is controllable by the agent,
we develop an instance of \emph{attentive dynamics model} (ADM) based on inverse dynamics $f_\text{inv}$.
The model
takes two consecutive input frames (observations) $s_{t-1}, s_{t} \in \mathcal{S}$ as input
and aims to predict the action ($a_{t-1} \in \mathcal A$) taken by the agent
to transition from $s_{t-1}$ to $s_{t}$:
\begin{align}
    \widehat{a}_{t-1} = f_\text{inv}(s_{t-1}, s_{t}).
\end{align}
Our key intuition is that the inverse dynamics model should
attend to the most relevant part of the observation,
which is controllable by the agent,
to be able to classify the actions.
We determine
whether %
each region in a $H \times W$ grid is controllable, or in other words,
useful for predicting the agent's action,
by using a spatial \emph{attention mechanism}
\citep{Bahdanau:ICLR2015:Attention,Xu:ICML2015:ShowAttendTell}.
An overview of the model is shown in Figure \ref{fig:figure1}.

\textbf{Model.}
To perform action classification, we first compute a convolutional feature map $\phi_t^s = \phi(s_t) \in \R^{H \times W \times K}$
based on the observation $s_t$ using a convolutional neural network $\phi$.
We estimate a set of {\em logit} (score) vectors, denoted $e_t (i, j) \in \mathbb{R}^{|\mathcal A|}$,
for action classification from each grid cell $(i, j)$ of the convolutional feature map.
The local convolution features and feature differences for consecutive frames are fed into a 
shared multi-layer perceptron (MLP) to derive the logits as:
\vspace*{-2pt}%
\begin{align}
    e_t(i, j) =
        \mathrm{MLP}\Big(\big[
            \phi^s_{t}(i, j) - \phi^s_{t-1}(i, j)  ;~  \phi^s_{t}(i, j)
    \big]\Big)
    \in \mathbb{R}^{|\mathcal A|}%
    .
\end{align}
\vspace*{-10pt}%
We then compute %
an attention mask $\alpha_t \in \mathbb R^{H \times W}$ corresponding to frame $t$, which
indicates the controllable parts of the observation $s_t$.
Such attention masks are computed via a separate MLP from the features of each region $(i, j)$,
and then converted into a probability distribution using softmax or sparsemax operators~\citep{Martins:2016:Sparsemax}:
\begin{align}
    \alpha_t = \mathrm{sparsemax}(\widetilde{\alpha}_t)
    \text{~~~~where~~~}
    \widetilde{\alpha}_t(i, j) = \mathrm{MLP}\big( \phi^s_t (i,j) \big),
\end{align}
so that $\sum_{i,j} \alpha_t(i,j)=1$.
The sparsemax operator is similar to softmax but yields a sparse attention, leading to more stable performance.
Finally, the logits $e_t(i, j)$ from all regions
are linearly combined using the attention probabilities $\alpha_t$:
\begin{align}
    p(\widehat a_{t-1} \mid s_{t-1}, s_{t}) = \softmax%
    \hspace*{-2pt}\left( {
            \textstyle \sum_{i,j} \alpha_{t}(i,j) \cdot e_t(i, j)
    } \right)%
    \in \mathbb R^{|\mathcal A|}%
    .
\end{align}

\vspace*{-6pt}
\textbf{Training.}
The model can be optimized with
the standard cross-entropy loss $\mathcal L_\text{action}(a^*_{t-1}, \widehat a_{t-1})$
with respect to the ground-truth action $a^*_{t-1} \in \mathcal{A}$
that the agent actually has taken.
Based on this formulation,
the attention probability $\alpha_t(i, j)$
should be high only on regions $(i, j)$ that are predictive of the agent's actions.
Our formulation enables learning to localize controllable entities %
in a self-supervised way without any additional supervisory signal, unlike some prior work (\eg,~\citep{Bellemare:AAAI2012:Contingency})
that adopts simulators to collect extra supervisory labels.

Optimizing the parameters of \ADM{} on on-policy data is challenging for several reasons.
First, the ground-truth action may be unpredictable for given pairs of frames, leading to noisy labels.
For example, actions taken in uncontrollable situations do not have any effect
(\eg, when the agent is in the middle of jumping in \MontezumaRevenge).
Second, since we train the \ADM{} online along with the policy, %
the training examples are not independently and identically distributed, and the data distribution can shift dramatically over time.
Third, the action distribution from the agent's policy can run into a low entropy%
\footnote{We note that an entropy regularization term (\eg, Eq.(\ref{eq:a2c_entropy})) is used when learning the policy.}%
, being biased towards certain actions. %
These issues may prevent the \ADM{} from
generalization to novel observations, which hurts exploration. %
Generally, we prefer models %
that quickly adapt to the policy and learn to localize the controllable regions in a robust manner.

To mitigate the aforementioned issues, we adopt a few additional objective functions. %
We encourage the attention distribution to attain a high entropy
by including an \emph{attention entropy regularization loss}, \ie, $\mathcal L_\text{ent} = -\mathcal H(\alpha_t)$.
This term penalizes over-confident attention masks, making the attention closer to uniform
whenever action prediction is not possible.
We also train the logits corresponding to each grid cell independently using a separate cross-entropy loss:
$p(\widehat a^{i,j}_{t-1} \mid e_t(i,j)) = \mathrm{softmax}(e_t(i,j))$.
These additional cross-entropy losses, denoted $\mathcal L_\text{cell}^{i,j}$, %
allow the model to learn from unseen %
observations even when attention fails to perform well at first.
The entire training objective becomes:
\begin{align}
    \label{eq:loss_ADM}
    \mathcal L^\text{ADM} ~=~ \mathcal L_\text{action} + \textstyle\sum_{i,j} \mathcal L_\text{cell}^{i,j} + \lambda_\text{ent} \mathcal L_\text{ent} %
\end{align}
where $\lambda_\text{ent}$ is a mixing hyperparameter.
\vspace*{-4pt}
\subsection{Count-based Exploration with Contingent Regions}
\vspace*{-5pt}

One natural way to take advantage of discovered contingent regions for exploration
is count-based exploration.
The ADM %
can be used to localize the controllable entity (\eg, the agent's avatar) from an observation $s_t$ experienced by the agent.
In 2D environments,
a natural discretization %
$
(x, y) = \textstyle\argmax_{(j, i)}  \alpha_t (i, j)
$
provides a good approximation of the agent's location within the current observation%
\footnote{%
To obtain more accurate %
localization by taking temporal correlation into account,
we can use exponential smoothing as
$\overline{\alpha}_t(i, j) = (1 - \omega_t) \overline{\alpha}_{t-1}(i, j) + \omega_t \alpha_t(i, j)$,
where ${\omega_t} = \max_{(i,j)}\{\alpha_t (i,j)\}$. %
}.
This provides a key piece of information about the current state of the agent.

\label{sec:overview}
Inspired by 
previous work \citep{Bellemare:NIPS2016:UnifyingCount,Tang:NIPS2017:SimHash},
we add an exploration bonus of $r^+$ to the environment reward, %
where    $r^+(s) = 1  / {\sqrt{\#( \psi(s) )}}$ and
$\#(\psi(s))$ denotes the visitation count of the (discrete) mapped state $\psi(s)$, 
{which consists of the contingent region $(x, y)$.
We want to find a policy $\pi$ that maximizes 
the expected discounted sum of environment rewards $r^\text{ext}$ %
plus count-based exploration rewards $r^+$, denoted
$%
\mathcal R = \mathbb{E_\pi}\big[\textstyle \sum_{t} \gamma^t \left(\beta_1 r^\text{ext}(s_t, a_t) + \beta_2 r^+(s_t)\right) \big]
$%
, where $\beta_1, \beta_2 \ge 0$ are hyperparameters
that balance the weight of environment reward and exploration bonus.
For every state $s_t$ encountered at time step $t$, we increase the counter value $\#(\psi(s_t))$ by 1 during training.
The full procedure is summarized in Algorithm \ref{alg:exp} in Appendix \ref{sec:summary_algorithm}.

\medskip

%% file: 04_experiments.tex
\vspace*{-5pt}
\section{Experiments}
\label{sec:experiments}
\vspace*{-5pt}

In the experiments below we investigate the following key questions:

\vspace*{-8pt}
\begin{itemize}[leftmargin=7mm]
    \setlength{\itemsep}{0pt}\setlength{\parskip}{0pt}
    \item Does the contingency awareness in terms of self-localization provide a useful state abstraction for exploration?
    \item How well can the self-supervised model discover the ground-truth abstract states?
    \item How well does the proposed exploration strategy perform against other exploration methods?
\end{itemize}
\vspace*{-5pt}
\begin{figure*}[tb]
\begin{center}
    \centerfloat
    \vspace*{-5pt}
    \includegraphics[width=1.03\linewidth]{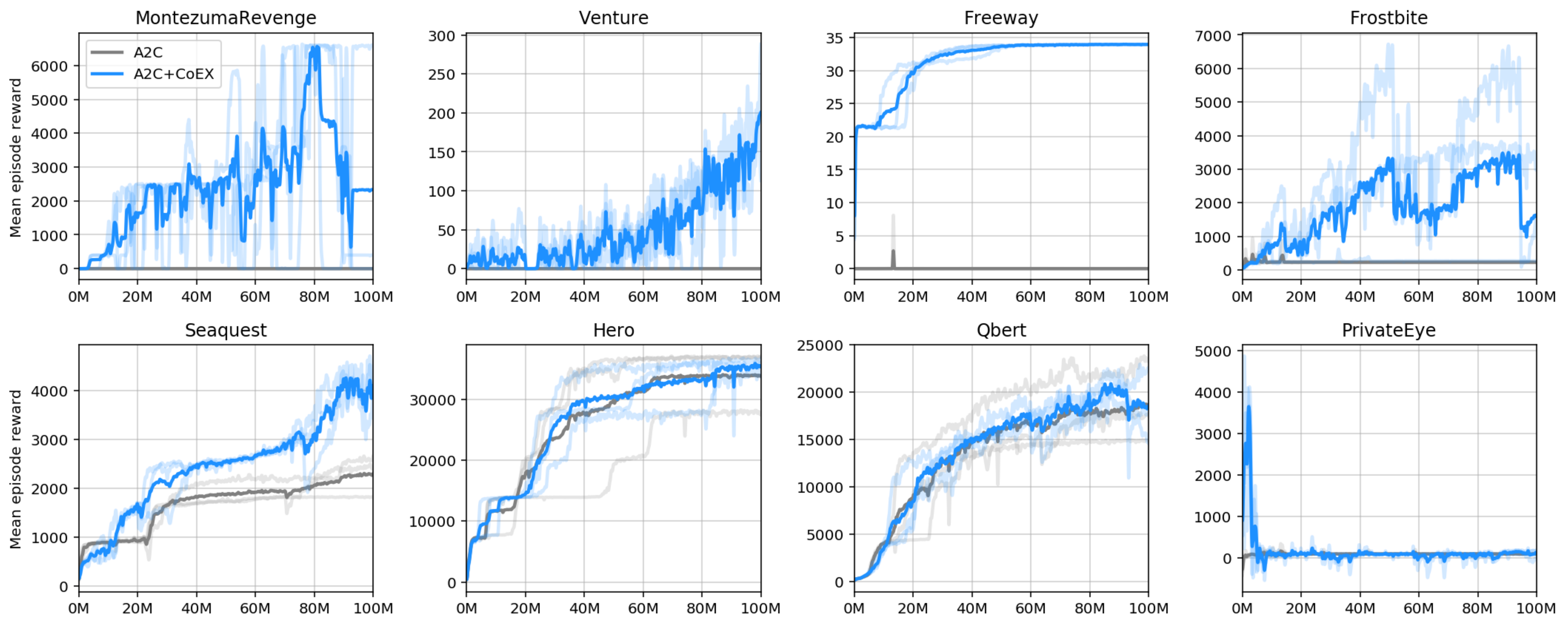}   %
    \vspace*{-5pt}
    \caption{
        Learning curves on several Atari games: A2C+\coex{} and A2C. %
        The x-axis represents total environment steps and
        the y-axis the mean episode reward averaged over 40 recent episodes.
        The mean curve is obtained by averaging over 3 random seeds, each shown in a light color.
    }
    \vspace*{-8pt}
    \label{fig:learning_curve}
\end{center} \end{figure*}

\vspace*{-4pt}
\subsection{Experiments with A2C}
\vspace*{-3pt}

We evaluate the proposed exploration strategy on several difficult exploration Atari 2600 games
from the Arcade Learning Environment (ALE) \citep{Bellemare:JAIR2013:ALE}.
We focus on 8 Atari games %
including \Freeway, \Frostbite, \Hero, \PrivateEye, \MontezumaRevenge, \Qbert, \Seaquest, and \Venture.
In these games, an agent without an effective exploration strategy
can often converge to a suboptimal policy.
For example, as depicted in Figure \ref{fig:learning_curve}, the Advantage Actor-Critic (A2C) baseline~\citep{Mnih:ICML2016:A3C}
achieves a reward close to $0$ on \MontezumaRevenge, \Venture, \Freeway, \Frostbite, and \PrivateEye, even after
$100$M steps of training. By contrast, our proposed technique, which augments A2C with
count-based exploration with the location information learned by the attentive dynamics model, denoted {\bf A2C+\coex{}}
(\coex{} stands for ``Contingency-aware Exploration''),
\iftrue %
    significantly outperforms the A2C baseline on six out of the 8 games.
\else %
significantly outperforms the A2C baseline on $6$ out of the $8$ games, %
presenting a new state-of-the-art for exploration in difficult Atari games (see Table \ref{tbl:exp_performance}).
\fi

We compare our proposed A2C+\coex{} technique against the following baselines:\footnote{%
In Section~\ref{sec:ppo}, we also report experiments using Proximal Policy Optimization~(PPO)~\citep{schulman2017:proximal} as a baseline, where our {\bf PPO+\coex{}} achieves the average score of >11,000 on \MontezumaRevenge.}

\begin{itemize}[leftmargin=7mm]
    \setlength{\itemsep}{0pt}\setlength{\parskip}{1pt}
 \item {\bf A2C}: an implementation adopted from OpenAI baselines~\citep{Dhariwal:2017:baselines} using the default hyperparameters,
 which serves as the building block of our more complicated baselines.
 \item {\bf A2C+Pixel-SimHash}: Following \citep{Tang:NIPS2017:SimHash}, we map 52$\times$52 gray-scale observations to
 $128$-bit binary codes using random projection followed by quantization~\citep{charikar2002}. Then, we add a count-based exploration bonus
 based on quantized observations.
\end{itemize}
\vspace*{-4pt}
As a control experiment, we evaluate {\bf A2C+\coexRAM{}$^*$}, our contingency-aware exploration method {together} with the ground-truth location information {obtained} from game's RAM. It is roughly an upper-bound of the performance of our approach.
\input{04_z-table1.tex}
\vspace*{-3pt}
\subsection{Implementation Details}
\vspace*{-3pt}
For the A2C \citep{Mnih:ICML2016:A3C} algorithm,
we use 16 parallel actors to collect the agent's experience, with 5-step rollout, which yields a minibatch of size 80 for on-policy transitions.
We use the last 4 observation frames stacked as input, each of which is resized to $84 \times 84$ and converted to grayscale
as in \citep{Mnih:Nature2015:DQN,Mnih:ICML2016:A3C}.
We set the end of an episode to when the game ends, rather than when the agent loses a life.
Each episode is initialized with a random number of no-ops \citep{Mnih:Nature2015:DQN}.
More implementation details can be found in Appendix \ref{sec:summary_algorithm} and \ref{sec:hyperparam_details}.

For the ADM,
we take observation frames of size $160 \times 160$ as input
(resized from the raw observation of size $210\times 160$).\footnote{%
    In some games such as Venture, the agent is depicted in very small pixels,
    which might be hardly recognizable in rescaled $84 \times 84$ images.
}
We employ a 4-layer convolutional neural network that produces a feature map $\phi(s_t)$
with a spatial grid size of $H \times W = 9 \times 9$.
As a result, the prediction of location coordinates lies in the $9 \times 9$ grid.

In some environments, the contingent regions within the visual observation alone are not sufficient to determine the exact location of the agent within the game;
for example, the coordinate cannot solely %
distinguish between different rooms in
\Hero, \MontezumaRevenge, and \PrivateEye, etc.
Therefore, we introduce a discrete context representation $c \in \mathbb{Z}$ %
that summarizes the high-level visual context in which the agent currently lies.
We use a simple clustering method similar to \citep{Kulis:ICML2012:BNPCluster},
which we refer to as \emph{observation embedding clustering}
that clusters the random projection vectors of the input frames
as in \citep{Tang:NIPS2017:SimHash},
so that different contexts are assigned to different clusters.
We further explain this heuristic approach more in detail in Appendix \ref{sec:context_embedding}.

In sparse-reward problems, the act of collecting a reward is rare but frequently instrumental for the future states of the environment.
The cumulative reward  $R_t = \sum_{t'=0}^{t-1} r^\text{ext}(s_{t'}, a_{t'})$
from the beginning of the episode up to the current step $t$, can provide a useful high-level  \emph{behavioral context} because collecting rewards can trigger significant changes to the agent's state
and as a result the optimal behavior can change as well. In this sense, the agent should revisit the previously visited location for exploration when the context changes.
For example, in \MontezumaRevenge, if the agent is in the first room and the cumulative reward is 0, %
we know the agent has not picked up the key and the optimal policy is to reach the key.
However, if the cumulative reward in the first room is 100, it means the agent has picked up the key and the next optimal goal is to open a door and move on to the next room.
Therefore, we could include the cumulative reward as a part of state abstraction for exploration, which leads to empirically better performance.

To sum up, for the purpose of count-based exploration,
we utilize
the location
$(x, y)$ of the controllable entity (\ie, the agent) %
in the current observation discovered by ADM (Section \ref{sec:dynamics}),
a context representation $c \in \mathbb{Z}$ that denotes the high level visual context,
and a cumulative environment reward $R \in \mathbb{Z}$ that represents the exploration behavioral state.
In such setting, we may denote $\psi(s) = (x, y, c, R)$.

\vspace*{-3pt}
\subsection{Performance of Count-Based Exploration}
\vspace*{-5pt}

Figure~\ref{fig:learning_curve}
shows the learning curves of the proposed methods %
on 8 Atari games.
The performance of our method A2C+\coex{} and A2C+\coexRAM{} as well as
the baselines A2C and A2C+Pixel-SimHash are summarized in Table \ref{tbl:a2c_performance}.
In order to find a balance between the environment reward and the exploration bonus reward,
we perform a hyper-parameter search for the proper weight of
the environment reward $\beta_1$ and the exploration reward $\beta_2$ for A2C+\coexRAM{}, as well as for A2C+\coex.
The hyper-parameters for the two ended up being the same, which is consistent with our results.
For fair comparison, we also search for the proper weight of environment reward for A2C baseline.
The best hyper-parameters for each game are shown in Table \ref{tbl:hyper-a2c} in Appendix \ref{sec:hyperparam_details}.

Compared to the vanilla A2C,
the proposed exploration strategy improves the score on all the hard-exploration games.
As shown in Table \ref{tbl:a2c_performance},
provided %
the representation $(x,y,c,R)$ is perfect, A2C+\coexRAM{} achieves a significant improvement over A2C
by encouraging the agent to visit novel locations, and could nearly solve these hard exploration games %
as training goes on.

Furthermore, A2C+\coex{} using representations learned with our proposed attentive dynamics model and observation embedding clustering
also outperforms the A2C baseline. Especially on \Freeway, \Frostbite, \Hero, \MontezumaRevenge, \Qbert and \Seaquest, the performance is comparable with A2C+\coexRAM{},
demonstrating %
the usefulness of the contigency-awareness information discovered by \ADM{}.
\begin{figure*}[t] \begin{center}
    \vspace*{-4pt}
    \includegraphics[width=\linewidth]{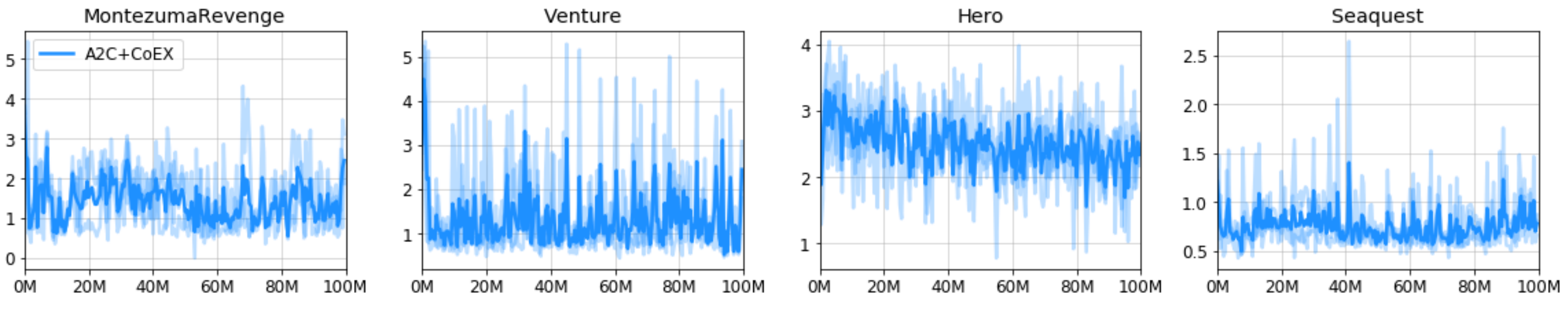}
    \vspace*{-15pt}
    \caption{
        Performance plot of ADM trained using on-policy samples from the A2C+\coex{} agent.
    }
    \label{fig:dynamics}
    \vspace*{-5pt}
    \vspace*{-2pt}
\end{center} \end{figure*}

\textbf{Comparison to other count-based exploration methods}.
Table~\ref{tbl:exp_performance} compares the proposed method with previous state-of-the-art results,
where our proposed method outperforms the other methods on 5 out of 8 games.
DQN-PixelCNN is the strongest alternative achieving a state-of-the-art performance on some of the most difficult sparse-reward games.
We argue that using Q-learning as the base learner with DQN-PixelCNN %
makes the direct comparison with A2C+\coex{} not completely adequate.
Note that the closest alternative count-based exploration method to A2C+\coex{} would be A3C+~\citep{Bellemare:NIPS2016:UnifyingCount},
which augments A3C \citep{Mnih:ICML2016:A3C} with exploration bonus derived from pseudo-count,
because A2C and A3C share a similar policy learning method.
With that in mind, one can observe a clear improvement of A2C+\coex{} over A3C+ on all of the 8 Atari games.

\vspace*{-3pt}
\subsection{Analysis of Attentive Dynamics Model}
\vspace*{-4pt}
\label{sec:analysis_inversedynamics}

We also analyze the performance of the ADM that learns
the controllable dynamics of the environment. %
As a performance metric, we report the average distance
between the ground-truth agent location $(x^*, y^*)$
and the predicted location $(x, y)$ within the $9 \times 9$ grid:
$\|(x, y) - (x^*, y^*)\|_2$.
The ground-truth location of the agent is extracted from RAM%
\footnote{%
Please note that the location from RAM is used only for analysis and evaluation purposes.},
then rescaled so that the observation image frame fits into the $9 \times 9$ grid. %

Figure~\ref{fig:dynamics} shows the results on 4 Atari games
(\MontezumaRevenge, \Seaquest, \Hero, and \Venture).
The ADM is able to quickly capture the location of the agent
without any supervision of localization, despite the agent constantly visiting new places.
Typically the predicted location is
on average 1 or 2 grid cells away from the ground-truth location.
Whenever a novel scene is encountered (\eg, the second room in \MontezumaRevenge at around 10M steps),
the average distance temporarily increases but quickly drops again as the model learns the new room.
We provide videos of the agents playing and localization information as the supplementary material.%
\footnote{%
    A demo video of the learnt policy and localization is available at
    {\footnotesize \url{\ProjectURL}}.
}

\begin{figure*}[tb] \begin{center}
    \vspace*{-4pt}
    \includegraphics[width=1.0\linewidth]{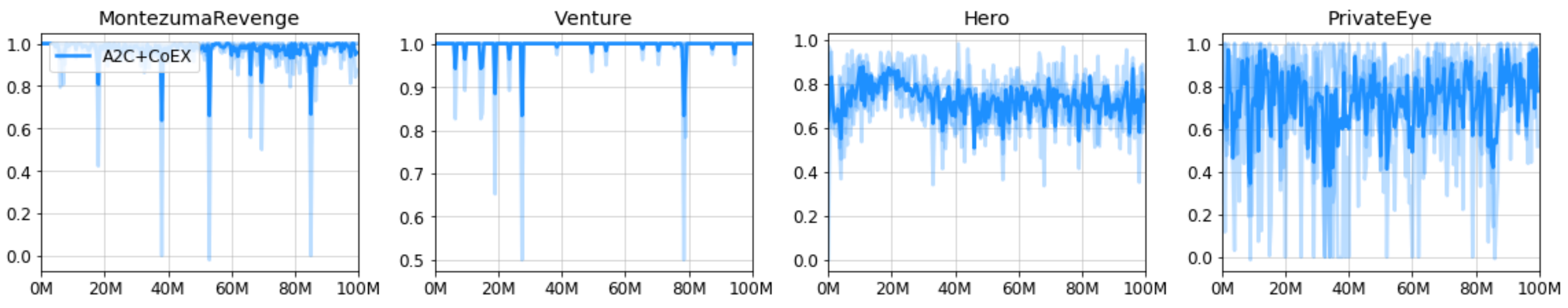}
    \vspace*{-15pt}
    \caption{
         Curves of ARI score %
         during training of A2C+\coex{},
         averaged over 100 recent observations.
    }
    \label{fig:ari_curve}
    \vspace*{-5pt}
    \vspace*{-3pt}
\end{center}
\end{figure*}

\vspace*{-3pt}
\subsection{Analysis of Observation Embedding Clustering}
\vspace*{-4pt}

To make the agent aware of
a change in high-level visual context (\ie, rooms in Atari games)
in some games such as \MontezumaRevenge, \Venture, \Hero, and \PrivateEye,
we obtain a representation of the high-level context and use it for exploration.
The high-level visual contexts are different from each other (different layouts, objects, colors, etc.), so the embedding generated by a random projection is quite distinguishable
and the clustering is accurate and robust.

For evaluation, given an observation in Atari games,
we compare the discrete representation
(\ie, which cluster it is assigned to) based on the embedding from random projection
to the ground-truth room number extracted from RAM.
The Adjusted Rand Index (ARI) \citep{Rand:ARI} measures the similarity between these two data clusterings.
The ARI may only yield a value between 0 and 1, and is exactly 1 when the clusterings are identical.

The curves of the Adjusted Rand Index are shown in Figure \ref{fig:ari_curve}.
For \MontezumaRevenge and \Venture, the discrete representation as room number is roughly as good as the ground-truth.
For \Hero and \PrivateEye, %
since there are many rooms quite similar \iffalse with \fi {to one another,} %
it is more challenging to accurately cluster the embeddings.
The samples shown in Figure \ref{fig:cluster_samples} in Appendix \ref{sec:context_embedding}
show reasonable performances of the clustering method on all these games.

\vspace*{-3pt}
\subsection{Additional Experiments with PPO} %
\label{sec:ppo}
\vspace*{-5pt}

We also evaluate the proposed exploration algorithm on \MontezumaRevenge
using the sticky actions environment setup \citep{machado2017:revisiting}
identical to the setup found in %
\citep{ICLR2019:RND}.
In the sticky action setup, the agent randomly repeats the previous action with probability of 0.25,
preventing the algorithm from simply memorizing the correct sequence of actions and relying on determinism.
The agent is trained with Proximal Policy Optimization~(PPO)~\citep{schulman2017:proximal}
in conjunction with the proposed exploration method
using 128 parallel actors to collect the experience.
We used reward normalization and advantage normalization as in \citep{Burda:2018:Curiosity}.

\begin{figure}[t]
  \centerfloat
  \vspace*{-4pt}
    \includegraphics[width=1.03\linewidth]{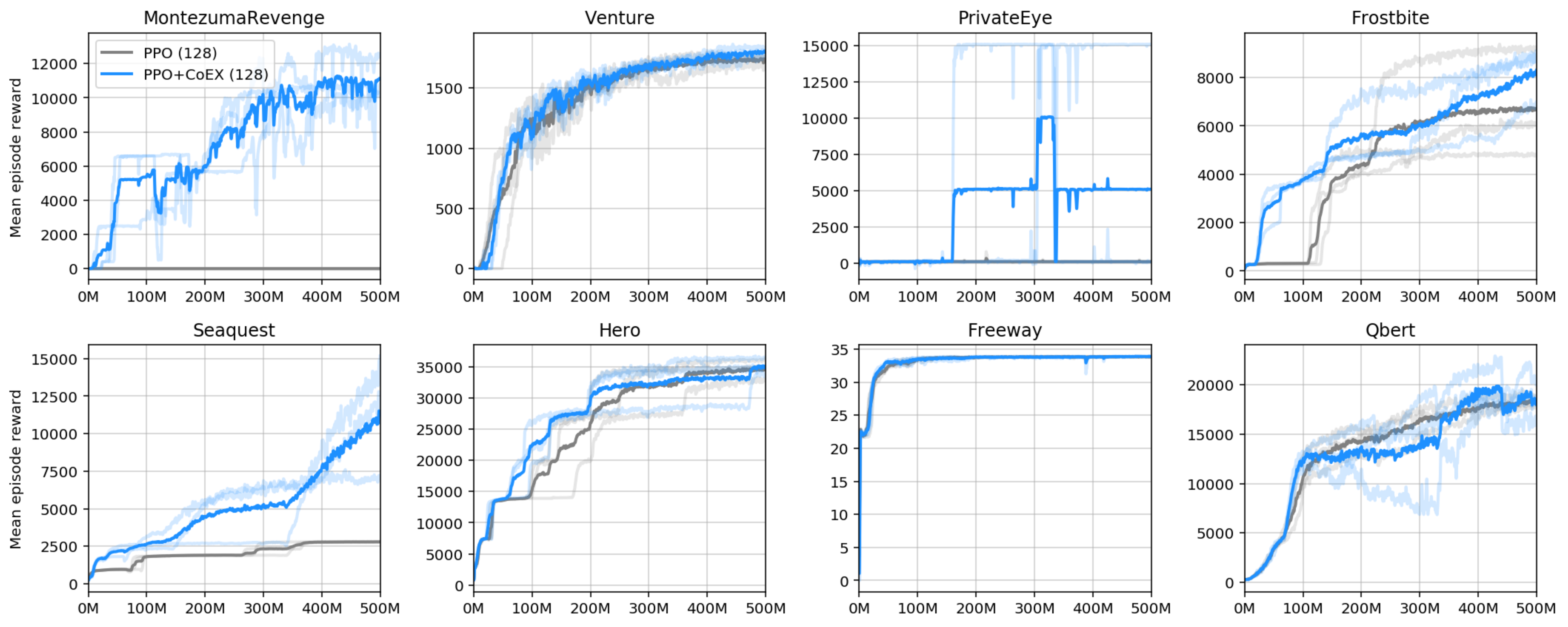}
  \vspace*{-7pt}
  \caption{The learning curve of PPO+\coex{} on several Atari games with sticky actions setup. %
  The x-axis represents the total number of environment steps and
  the y-axis the mean episode reward averaged over 40 recent episodes.
  The mean curve is obtained by averaging over 3 random seeds, each shown in a light color.}
  \label{fig:ppo_coex}
\end{figure}

The method, denoted {\bf PPO+\coex{}}, %
achieves the score of {11,618} %
at 500M environment steps (2 billion frames) on \MontezumaRevenge, when averaged over 3 runs.
The learning curve is illustrated in Figure \ref{fig:ppo_coex}.
\alert{%
}
Since the vanilla PPO baseline achieves a score near 0 (our runs) or 1,797%
~\citep{ICLR2019:RND},
this result is not solely due to the benefits of PPO.
There is another approach "Exploration by Random Network Distillation"
~\citep{ICLR2019:RND}
concurrent with our work which achieves similar performance by following a slightly different philosophy.

\input{04_z-table-ppo}

\vspace*{-2pt}
\subsection{Discussions and Future Work}
\vspace*{-5pt}

This paper investigates whether discovering controllable dynamics via
an attentive dynamics model (\ADM{})
can help exploration in challenging sparse-reward environments.
We demonstrate the effectiveness of this approach by achieving significant improvements
on notoriously difficult video games. That being said, %
we acknowledge that our approach has certain limitations.
Our currently presented instance of state abstraction method mainly focuses on
controllable dynamics and employs a simple clustering scheme to abstract away uncontrollable elements of the scene.
In more general setting, one can imagine using attentive (forward or inverse) dynamics models
to learn an effective and compact abstraction of the controllable and uncontrollable dynamics as well,
but we leave this to future work.

Key elements of the current \ADM{} method include the use of spatial attention and modelling of the dynamics.
These ideas can be generalized by a set of attention-based dynamics models (ADM) operating in forward, inverse, or combined mode.
Such models could use attention over a lower-dimensional embedding that corresponds to an intrinsic manifold structure from the environment
(\ie, intuitively speaking, this also corresponds to \emph{being self-aware of (e.g., locating) where the agent is in the abstract state space}). %
Our experiments with the inverse dynamics model suggest that the mechanism does not have to be perfectly precise, allowing for some error in practice. We speculate that mapping to such subspace could be obtained by techniques of embedding learning.
We note that RL environments with different visual characteristics may require different forms of attention-based techniques and properties of the model (\eg, partial observability).
Even though this paper focuses on 2D video games, we believe that the presented high-level ideas of learning contingency-awareness (with attention and dynamics models) are more general and could be applicable to more complex 3D environments with some extension. We leave this as future work.

%% file: 04_z-table1.tex
\newcommand{\B}[1]{\bf#1}
\newcommand{\RED}[1]{{\color{red}#1}}
\newcommand{\s}[1]{\small#1}                %
\def\pp{\hspace*{.18cm}}                %

\renewcommand{\arraystretch}{1.37}

\begin{table}[t]
    \tabcolsep=0.15cm
    \centering
    \vspace{-5pt}
\ifdefined\nipsstyle\vspace*{1pt}\fi
    \small
    \hspace*{-0.4cm}
    \begin{center}
    \begin{tabular}{@{}l|rrrrrrrr@{}}
        \hline
    \iffalse................................\fi Method & \s Freeway & \s Frostbite & \s Hero & \s Montezuma & \s PrivateEye & \s Qbert & \s Seaquest & \s Venture \\
        \hline
        A2C                                     & 7.2         & 1099         & 34352   & 13             & 574           & 19620    & 2401        & 0          \\
        A2C+Pixel-SimHash                       & 0.0        & 829          & 28181   & 412          & 276         & 18180    & 2177        & 31         \\
        A2C+\coex{}                                & \B 34.0    & \B 4260       &\B 36827   & \B 6635      & \B 5316          & \B 23962    &\B 5169        & \B 204        \\
        \hline
        A2C+\coexRAM{}$^*$                      &  34.0    &  4418        & 36765   & 6600         &  24296      &24422  & 6113    &  1100    \\
        \hline
    \end{tabular}
    \end{center}

\ifdefined\nipsstyle\medskip\vspace*{2pt}\fi
    \vspace*{-9pt}
    \caption{
        Performance of our method and its baselines on Atari games: maximum mean scores (averaged over 40 recent episodes)
        achieved over total 100M environment timesteps (400M frames) of training, averaged over 3 seeds.
        The best entry in the group of experiments without supervision is shown in bold.
        $^*$ denotes that A2C+\coexRAM{} acts as a control experiment, which includes some supervision.
        More experimental results on A2C+\coexRAM{} are shown in Appendix \ref{sec:a2c_exp_ram}.
    }
    \label{tbl:a2c_performance}
    \vspace*{-5pt}
\end{table}

\begin{table}[t]
    \tabcolsep=0.15cm
    \centering
    \small
    \hspace*{-0.25cm}
    \begin{center}
    \begin{tabular}{@{} l @{} r @{\pp}|@{\pp} r @{\pp} r @{\pp} r @{\pp} r @{\pp} r @{\pp} r @{\pp} r @{\pp} r @{}}
        \hline
    \iffalse.......................\fi Method & \#Steps & \s Freeway & \s Frostbite & \s Hero   & \s Montezuma & \s PrivateEye & \s Qbert & \s Seaquest & \s Venture \\
    \hline
    A2C+\coex{} (Ours)                          &   50M   & 33.9       &  3900        &  31367  &   4100     & 5316          & 17724  &  2620      & 128        \\
    A2C+\coex{} (Ours)                          &   100M  & \B 34.0    &  4260      &\B 36827   & \B 6635      &  5316       & \B 23962    &\B 5169    &  204        \\
    \hline
        DDQN+                                   & 25M        & 29.2      & {-}      &  {20300}& 3439       & {1880}    & -        & -           & 369        \\
        A3C+                                    & 50M        & 27.3      & 507          & 15210     & 142          & 100           & 15805    & 2274        & 0          \\
        TRPO-AE-SimHash                         & 50M        & 33.5      & \B 5214      & -         & 75           & -             & -        & -           & 445        \\
        Sarsa-$\phi$-EB                         & 25M        & 0.0       & 2770         & -         & 2745         & -             & 4112     & -           & 1169       \\
        DQN-PixelCNN                            & 37.5M        & 31.7      & -            & -         & 2514         & \B 15806      & 5501     & -           & \B 1356    \\
        Curiosity-Driven                        & 25M        & 32.8      & -            & -         & 2505         & 3037          & -        & -           & 416   \\
        \hline
    \end{tabular}
    \end{center}
    \vspace*{-8pt}
    \caption{
        Performance of our method and state-of-the-art exploration methods on Atari games.
        For fair comparison, we report the maximum mean score achieved over the specific number of timesteps during training, averaged over 3 seeds.
        The best entry is shown in bold.
        Baselines (for reference) are:
        DDQN+ and A3C+ \citep{Bellemare:NIPS2016:UnifyingCount},   %
        TRPO-AE-SimHash \citep{Tang:NIPS2017:SimHash},
        Sarsa-$\phi$-EB \citep{Martin:1706.08090},
        DQN-PixelCNN \citep{Ostrovski:ICML2017:ExplorationDensity},
        and Curiosity-Driven \citep {Burda:2018:Curiosity}.
        The numbers for DDQN+ %
        were taken from \citep{Tang:NIPS2017:SimHash} or were read from a plot.
    }
    \label{tbl:exp_performance}
\vspace*{-5pt}
\end{table}

%% file: 04_z-table-ppo.tex
\renewcommand{\arraystretch}{1.38}

\begin{table}[t]
    \tabcolsep=0.14cm
    \centering
    \centerfloat
    \small
    \begin{tabular}{@{}lc|rrrrrrrr@{}}
        \hline
    \iffalse....................\fi Method & \#Steps   & \s Freeway & \s Frostbite & \s Hero & \s Montezuma & \s PrivateEye & \s Qbert & \s Seaquest & \s Venture \\
        \hline
        PPO                                & 500M      & \B 34.0   & 7340         & 36263   & 29           & 942           & 19980    & 2806        & 1875          \\
        PPO+\coex{}                        & 500M      & \B 34.0   & \B 9076      &\B 36664 & \B 11618     & \B 11000      & \B 22647 &\B 11794     & \B 1916        \\
        \hline
    \end{tabular}

    \vspace*{-5pt}
    \caption{
        Performance of PPO and PPO+\coex : maximum mean scores (average over 40 recent episodes)
        achieved over total 500M environment steps (2B frames) of training, averaged over 3 seeds.
    }
    \label{tbl:ppo_performance}
\end{table}

%% file: 05_conclusion.tex
\section{Conclusion}
\label{sec:conclusion}
We proposed a method of providing contingency-awareness through an attentive dynamics model (ADM).
It enables approximate self-localization for an RL agent in 2D environments (as a specific perspective).
The agent is able to estimate its position in the space and therefore benefits from a compact and informative representation of the world.
This idea combined with a variant of count-based exploration achieves strong results in various sparse-reward Atari games. Furthermore, we report state-of-the-art results of >11,000 points on the infamously challenging \MontezumaRevenge without using expert demonstrations or supervision.
Though in this work we focus mostly on 2D environments in the form of sparse-reward Atari games, we view our presented high-level concept and approach as a stepping stone towards more universal algorithms capable of similar abilities in various RL environments.

%% file: 10_appendix.tex
\ifdefined\workshopversion %
    \onecolumn

    \thispagestyle{empty}
\makeatletter
    {\center\baselineskip 18pt
                       \toptitlebar{\Large\bf 
    [Supplementary Material / Appendix] \\[5pt]
    \@title
           }\bottomtitlebar}
\makeatother
\fi

{\LARGE \textsc{Appendix}}\vspace{1em}

\appendix
\vspace*{-10pt}

\section{Summary of Training Algorithm}
\label{sec:summary_algorithm}

\begin{algorithm}
\begin{algorithmic}
\STATE Initialize parameter $\theta_\text{ADM}$ for attentive dynamics model $f_\text{ADM}$
\STATE Initialize parameter $\theta_\text{A2C}$ for actor-critic network
\STATE Initialize parameter $\theta_c$ for context embedding projector if applicable (which is not trainable)
\STATE Initialize transition buffer $\mathcal{E} \gets \emptyset$
\FOR{each iteration}
\STATE {\textit{$\vartriangleright$ Collect on-policy transition samples, distributed over $K$ parallel actors}}
	\FOR{each step $t$}
		\STATE $s_t \gets $ Observe state
		\STATE $a_t \sim \pi_\theta(a_t|s_t)$
        \STATE $s_{t+1},r^\text{ext}_t \gets$ Perform action $a_t$ in the environment
        \smallskip
        \STATE {\textit{$\vartriangleright$ Compute the contingent region information}}
        \STATE $\overline\alpha_{t+1} \leftarrow $ Compute the attention map of $s_{t+1}$ using $f_\text{ADM}$
        \STATE $c(s_{t+1}) \leftarrow$ Compute the observation embedding cluster of $s_{t+1}$ (Algorithm \ref{alg:clustering}) 
        \smallskip
        \STATE {\textit{$\vartriangleright$ Increment state visitation counter based on the representation}}
        \STATE $\psi(s_{t+1}) \gets$ ($\argmax_{(i,j)} \overline{\alpha}_{t+1}(i,j)$, $c(s_{t+1})$, $\lfloor \sum_{k=0}^t r_k^\text{ext} \rfloor$)
		\STATE $\#(\psi(s_{t+1})) \gets \#(\psi(s_{t+1}))+1$
		\STATE $r^{+}_t \gets \frac{1}{\sqrt{\#(\psi(s_{t+1}))}}$
        \smallskip
        \STATE Store transition $\mathcal{E} \gets \mathcal{E} \cup \left\{(s_t, a_t, s_{t+1}, \beta_1 \mathrm{clip}(r_t^\text{ext}, -1, 1)+\beta_2 r^{+}_t)\right\}$
	\ENDFOR
	\STATE {\textit{$\vartriangleright$ Perform actor-critic using on-policy samples in $\mathcal{E}$}}
    \STATE $\theta_\text{A2C} \gets \theta_\text{A2C} - \eta \nabla_{\theta_\text{A2C}} \mathcal{L}^\text{A2C}$
	\STATE {\textit{$\vartriangleright$ Train the attentive dynamics model using on-policy samples in $\mathcal{E}$}}
    \STATE $\theta_\text{ADM} \gets \theta_\text{ADM} - \eta \nabla_{\theta_\text{ADM}} \mathcal{L}^\text{ADM}$
\STATE Clear transition buffer $\mathcal{E} \gets \emptyset$
\ENDFOR
\end{algorithmic}
\caption{A2C+\coex{}}%
\label{alg:exp}
\end{algorithm}

The learning objective $\mathcal L^\text{ADM}$ is from Equation (\ref{eq:loss_ADM}).
The objective $\mathcal L^\text{A2C}$ of Advantage Actor-Critic (A2C) is as in \citep{Mnih:ICML2016:A3C,Dhariwal:2017:baselines}:
\begin{align}
    \label{eq:a2c_loss}
    \mathcal L^\text{A2C} &=
        \mathbb E_{(s,a,r) \sim \mathcal{E}} \bigg[
                \mathcal L^\text{A2C}_\text{policy} + \frac{1}{2} \mathcal L^\text{A2C}_\text{value}
            \bigg]
    \\
    \label{eq:a2c_policy}
    \mathcal L^\text{A2C}_\text{policy} &=
    - \log \pi_{\theta}(a_t | s_t) (R_t^{n} - V_\theta (s_t)) - \alpha \mathcal{H}_t (\pi_\theta)
    \\
    \label{eq:a2c_value}
    \mathcal L^\text{A2C}_\text{value} &=
    \frac{1}{2} \Big( V_\theta (s_t) - R_t^{n} \Big)^2
    \\
    \label{eq:a2c_entropy}
    \mathcal{H}_t (\pi_\theta) &=
    - \textstyle \sum_{a} \pi_\theta(a|s_t) \log \pi_\theta(a | s_t)
\end{align}
where $R_t^{n} = \sum_{i=0}^{n-1} \gamma^i r_{t+i} + \gamma^n V_\theta (s_{t+n})$ is the $n$-step bootstrapped return %
and $\alpha$ is a weight for the standard entropy regularization loss term $\mathcal H_t(\pi_\theta)$.
We omit the subscript as $\theta = \theta_\text{A2C}$ when it is clear.

\clearpage

\section{Architecture and Hyperparameter Details}
\label{sec:hyperparam_details}

The architecture details of the attentive dynamics model (ADM), the policy network, and hyper-parameters are as follows.

\renewcommand{\arraystretch}{1.4}

\begin{table}[H]
\small
\centering
\caption{Network architecture and hyperparameters}
    \vspace*{-5pt}
\smallskip
\label{tbl:hyper-archi}
\begin{tabular}{l l l l}
\toprule
\multicolumn{2}{c}{Hyperparameters} & Value \\
\midrule
\multicolumn{2}{l}{Policy and Value Network Architecture} &
~Input: 84x84x1 \\
                                      & & - Conv(32-8x8-4) & /ReLU\\
                                      & & - Conv(64-4x4-2) & /ReLU\\
                                      & & - Conv(64-3x3-1) & /ReLU \\
                                      & & - FC(512)        & /ReLU\\
                                      & & - FC($|\mathcal A|$), FC(1)\\
\midrule
\multicolumn{2}{l}{ADM Encoder Architecture} &
~Input: 160x160x3 \\
                         & & - Conv(8-4x4-2) & /LeakyReLU\\
                         & & - Conv(8-3x3-2) & /LeakyReLU\\
                         & & - Conv(16-3x3-2) & /LeakyReLU\\
                         & & - Conv(16-3x3-2) & /LeakyReLU\\
\multicolumn{2}{l}{ MLP Architecture for $e_t(i,j)$}
                         & FC(1296,256) &/ReLU \\
                                &&- FC(256,128)  &/ReLU \\
                                &&- FC(128,$|\mathcal A|$) \\
\multicolumn{2}{l}{MLP Architecture for $\widetilde{\alpha}_t(i,j)$}
                & FC(1296,64) &/ReLU \\
                                                 &&- FC(64,64) &/ReLU \\
                                                 &&- FC(64,1) \\
\multicolumn{2}{l}{$\lambda_\text{ent}$ for Loss}
                & 0.001 \\
\midrule
A2C & Discount Factor $\gamma$  & 0.99 \\
    & Learning Rate (RMSProp) & 0.0007 \\
    & Number of Parallel Environments & 16 \\
    & Number of Roll-out Steps per Iteration & 5 \\
    & Entropy Regularization of Policy ($\alpha$) & 0.01 \\
\midrule
PPO
    & Discount Factor $\gamma$  & 0.99 \\
    & $\lambda$ for GAE               & 0.95    \\
    & Learning rate (Adam)            & 0.00001 \\
    & Number of Parallel Environments & 128 \\
    & Rollout Length                  & 128    \\ 
    & Number of Minibatches           & 4      \\ 
    & Number of Optimization Epochs   & 4      \\ 
    & Coefficient of Extrinsic and Intrinsic reward & $\beta_1 = 2, \beta_2 = 1$      \\ 
    & Entropy Regularization of Policy ($\alpha$)           & 0.01   \\ 
\bottomrule
\end{tabular}
    \vspace*{-5pt}
\end{table}

\clearpage

\begin{table}[H]
\small
\centering
\caption{The list of hyperparameters used for A2C+\coex{} in each game.
    For the four games where there is no change of high-level visual context
    (\Freeway, \Frostbite, \Qbert and \Seaquest),
    we do not include $c$ in the state representation $\psi(s)$, hence there is no $\tau$.
    The same values of $\tau$ are used in PPO+\coex{}.
}
\label{tbl:hyper-a2c}
\begin{tabular}{l l l l l}
\toprule
Games   & $\beta_1$ in A2C+\coex{}  & $\beta_2$ in A2C+\coex{} &$\beta_1$ in A2C & $\tau$ for clustering\\
\midrule
\Freeway & 10& 10 & 10 &-\\
\Frostbite & 10 & 10 & 10&- \\
\Hero &1 &0.1 & 1&0.7 \\
\MontezumaRevenge &10 &10 &10 &0.7\\
\PrivateEye &10 &10 &10 & 0.55\\
\Qbert &1 &0.5 &1 &-\\
\Seaquest &1 &0.5 &10 &-\\
\Venture &10 &10 & 10&0.7\\
\bottomrule
\end{tabular}
    \vspace*{-5pt}
\end{table}

\renewcommand{\arraystretch}{1.4}  %

\clearpage

\section{Experiment with RAM Information}
\label{sec:a2c_exp_ram}
\begin{figure*}[bt] \begin{center}
    \centerfloat
    \includegraphics[width=1.03\linewidth]{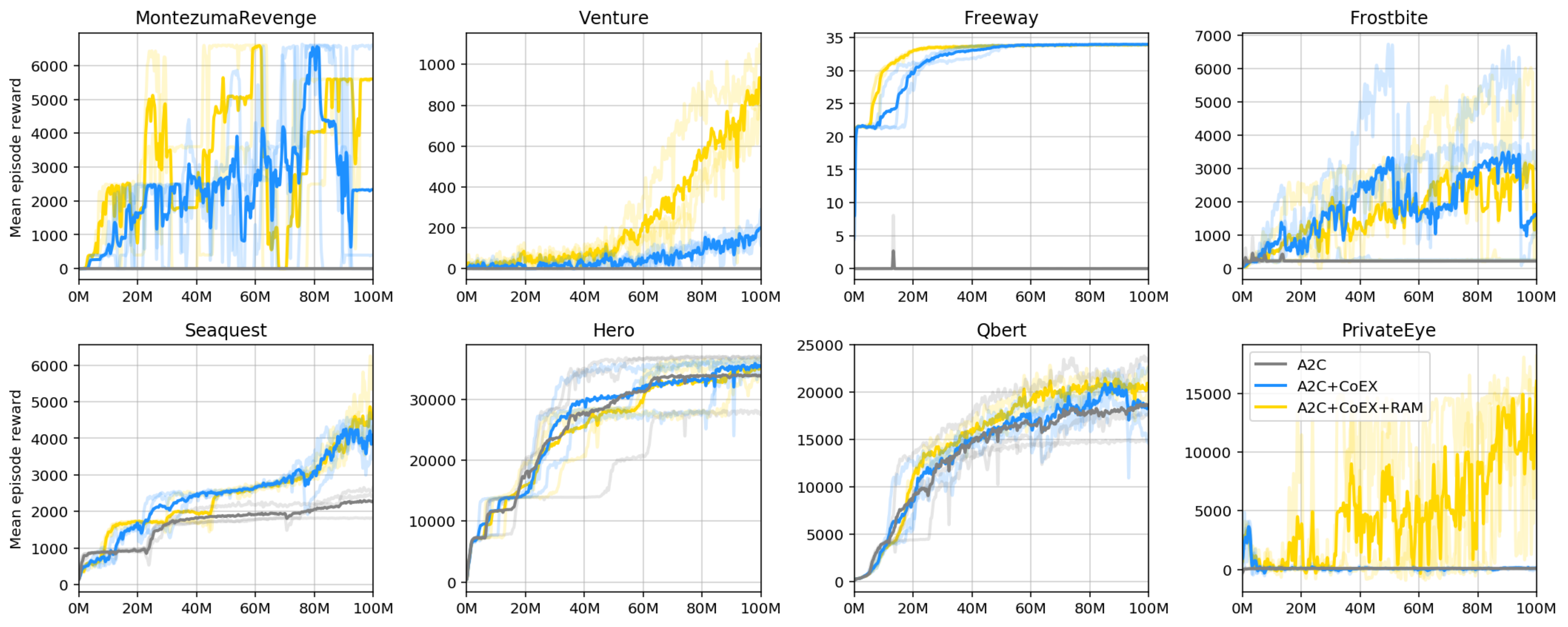}
    \vspace*{-5pt}
    \caption{
        Learning curves on several Atari games: A2C, A2C+\coex{}, and A2C+\coexRAM{}.
    }
    \label{fig:a2c_exp_ram}
    \vspace*{-5pt}
\end{center} \end{figure*}

In order to understand the performance of exploration with perfect representation,
we extract the ground-truth location of the agent and the room number from RAM,
and then run count-based exploration with the perfect $(x,y,c,R)$. Figure \ref{fig:a2c_exp_ram} shows the learning curves of the experiments;
we could see A2C+\coexRAM{} acts as an upper bound performance of our proposed method.

\section{Observation Embedding Clustering}
\label{sec:context_embedding}
We describe the detail of a method to obtain the observation embedding.
Given an %
observation of shape $(84, 84, 3)$, we flatten the observation and project it to an embedding of dimension $64$.
We randomly initialize the parameter of the fully-connected layer for projection, and keep the values unchanged during the training to make the embedding stationary.

For the embedding of these observations, we cluster them based on a threshold value $\tau$.
The value of $\tau$ for each game with change of rooms is listed in Table \ref{tbl:hyper-a2c}.
If the distance between the current embedding and the center $\mathrm{mean}(c)$ of a cluster $c$ is less than the threshold,
we assign this embedding to %
the cluster with the smallest distance and update its center with the mean value of all embeddings belonging to this cluster.
If the distance between the current embedding and the center of any cluster is larger than the threshold,
we create a new cluster and this embedding is assigned to this new cluster.

\begin{algorithm}
\begin{algorithmic}
\STATE Initialize parameter $\theta_c$ for context embedding projector if applicable (which is not trainable)
\STATE Initialize threshold $\tau$ for clustering
\STATE Initialize clusters set $C \gets  \emptyset$
\FOR{each observation $s$}
\STATE {\textit{$\vartriangleright$ Get embedding of the observation from the random projection}}
\STATE $v \gets f_{\theta_c}(s)$
\STATE {\textit{$\vartriangleright$ Find a cluster to which the current embedding fits, if any}}
\STATE Find a cluster $c \in C$ with smallest $\| \mathrm{mean}(c) - v \| \le \tau$, or $\mathrm{NIL}$ if there is no such
	\IF{$c \neq \mathrm{NIL}$}
	    \STATE $c \gets c \cup v$
	\ELSE
	    \STATE {\textit{$\vartriangleright$ if there's no existing cluster that $v$ should be assigned to, create a new one}}
	    \STATE $C \gets C \cup \{ v \}$
	\ENDIF
\ENDFOR
\end{algorithmic}
\caption{Observation Embedding Clustering}
\label{alg:clustering}
\end{algorithm}

\begin{figure*}[bt] \begin{center}
    \includegraphics[width=0.5555\linewidth]{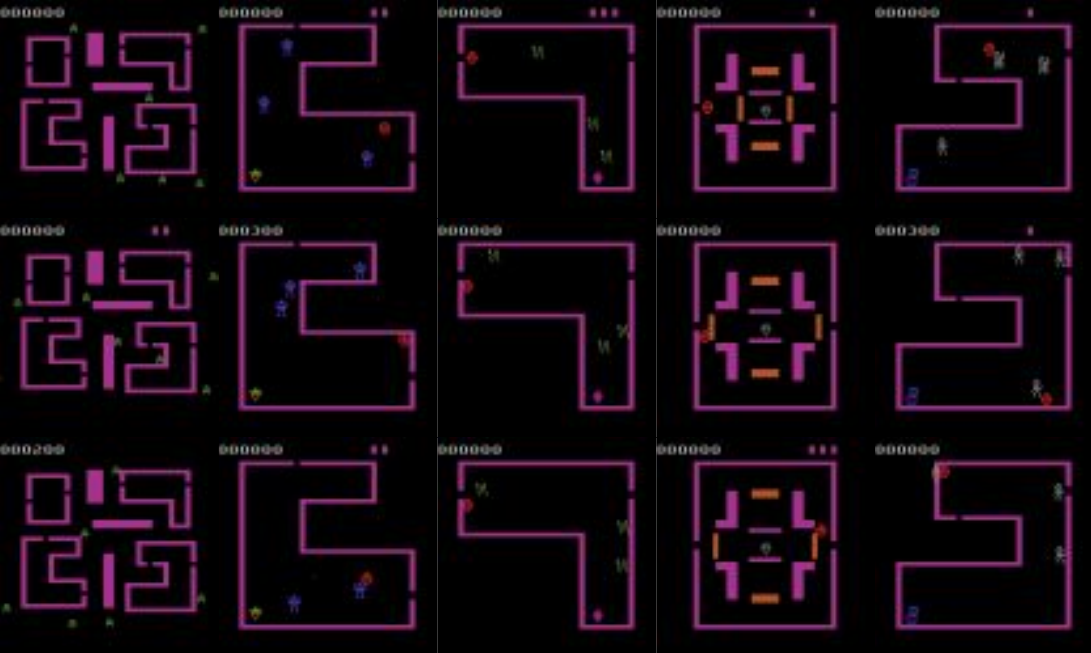}
    \vspace*{-0.3cm}
    \includegraphics[width=\linewidth]{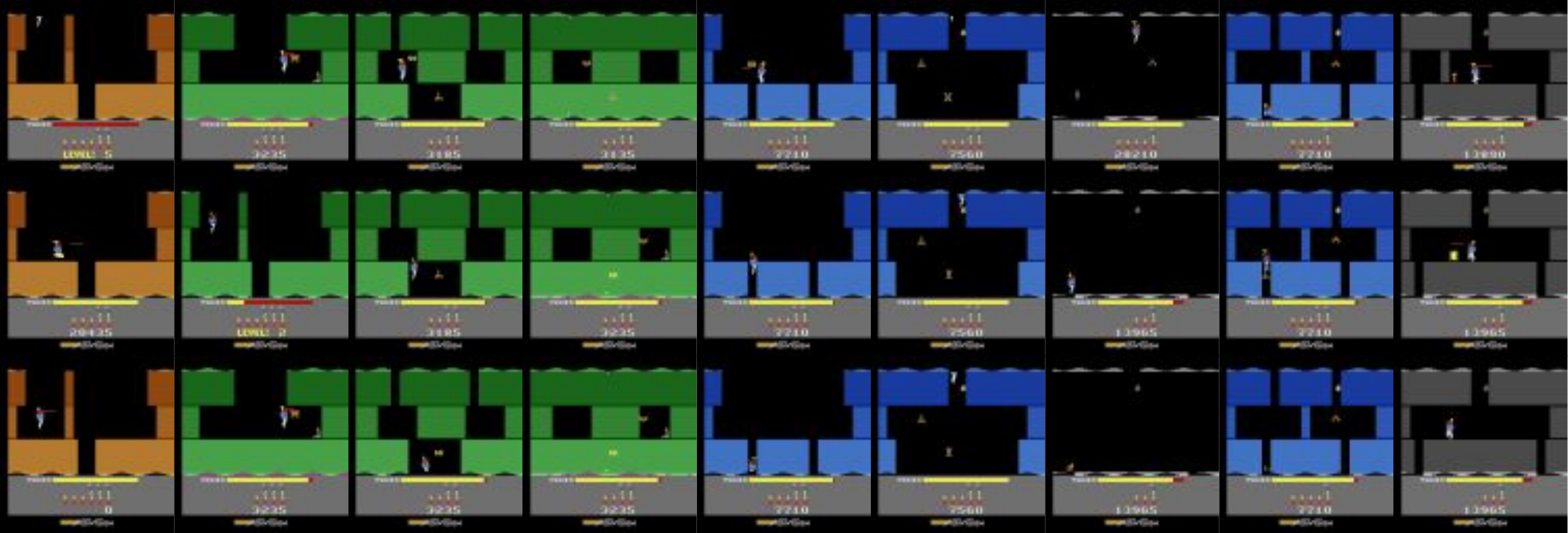}
    \vspace*{-0.3cm}
    \includegraphics[width=\linewidth]{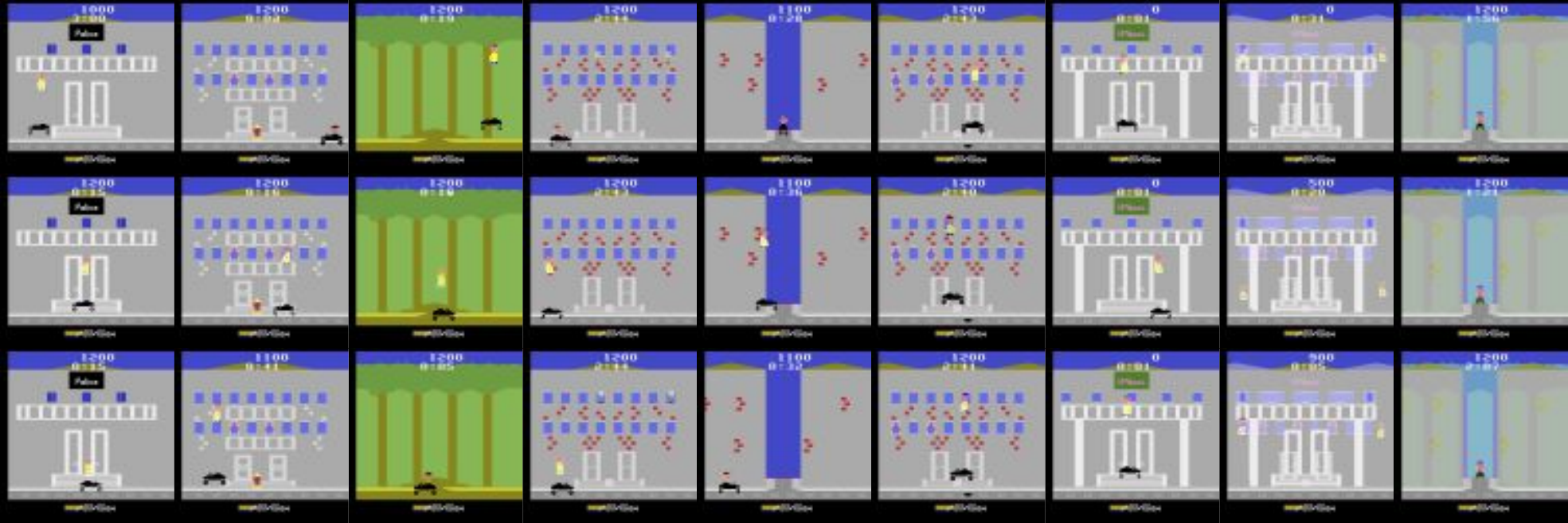}
    \vspace*{-0.3cm}
    \includegraphics[width=\linewidth]{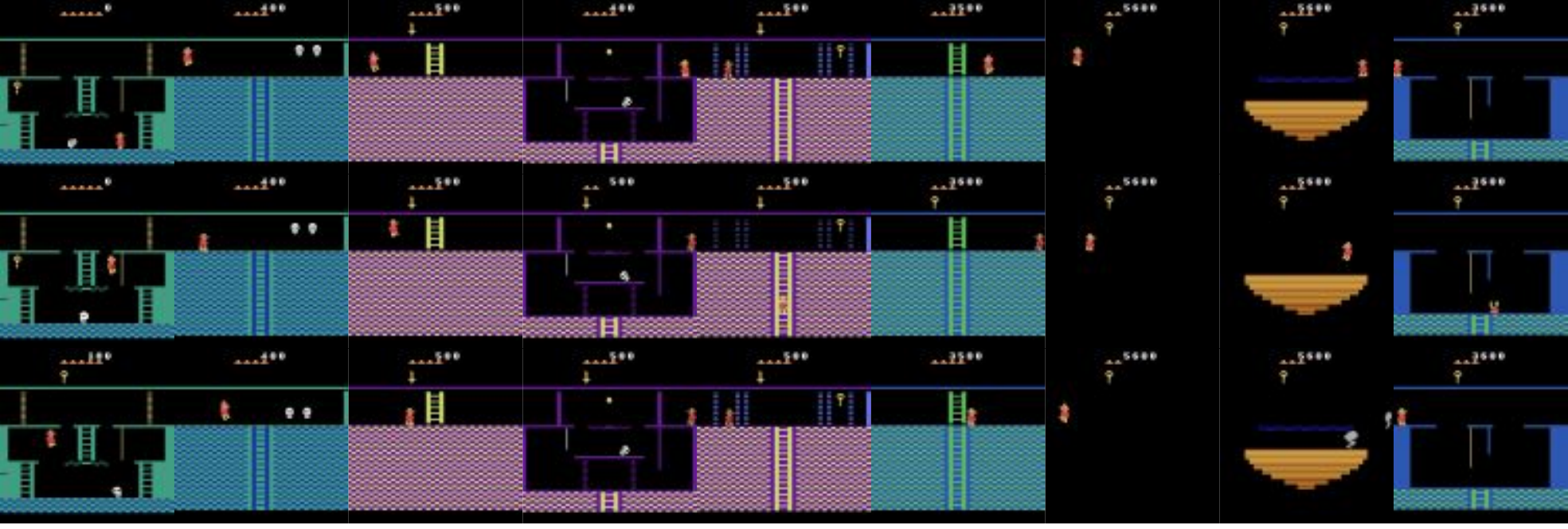}
    
    \vspace*{0.3cm}
    \caption{
        Sample of clustering results for \Venture, \Hero, \PrivateEye, and \MontezumaRevenge. Each column is one cluster, and we show 3 random samples assigned into this cluster.
    }
    \label{fig:cluster_samples}
\end{center} \end{figure*}

In Figure \ref{fig:cluster_samples}, we also show the samples of observation in each cluster.
We could see observations from the same room are assigned to the same cluster and different clusters correspond to different rooms.

\clearpage

%% file: 11_additional.tex
\clearpage

\FloatBarrier

\section{Ablation Study on Attentive Dynamics Model}
\label{sec:ablation_adm}

We conduct a simple ablation study on the learning objectives of ADM, described in Equation (\ref{eq:loss_ADM}).
We evaluate the performance of ADM when trained on the same trajectory data under different combinations of loss terms,
simulating batches of on-policy transition data to be replayed.
The sample trajectory was obtained from an instance of A2C+CoEX+RAM and kept same across all the runs,
which allows a fair comparison between different variants.
We compare the following four methods:
\begin{itemize}[leftmargin=7mm]
\setlength{\itemsep}{0pt}\setlength{\parskip}{3pt}
\item ADM (action)             : train ADM using $\mathcal L_\text{action}$ only
\item ADM (action, cell)       : train ADM using $\mathcal L_\text{action}$ and $\mathcal L_\text{cell}$
\item ADM (action, ent)        : train ADM using $\mathcal L_\text{action}$ and $\mathcal L_\text{ent}$
\item ADM (action, cell, ent)  : train ADM using all losses ($\mathcal L_\text{action}, \mathcal L_\text{cell}, \mathcal L_\text{ent}$)
\end{itemize}

\begin{figure*}[th]
\begin{center}
    \includegraphics[width=0.99\linewidth]{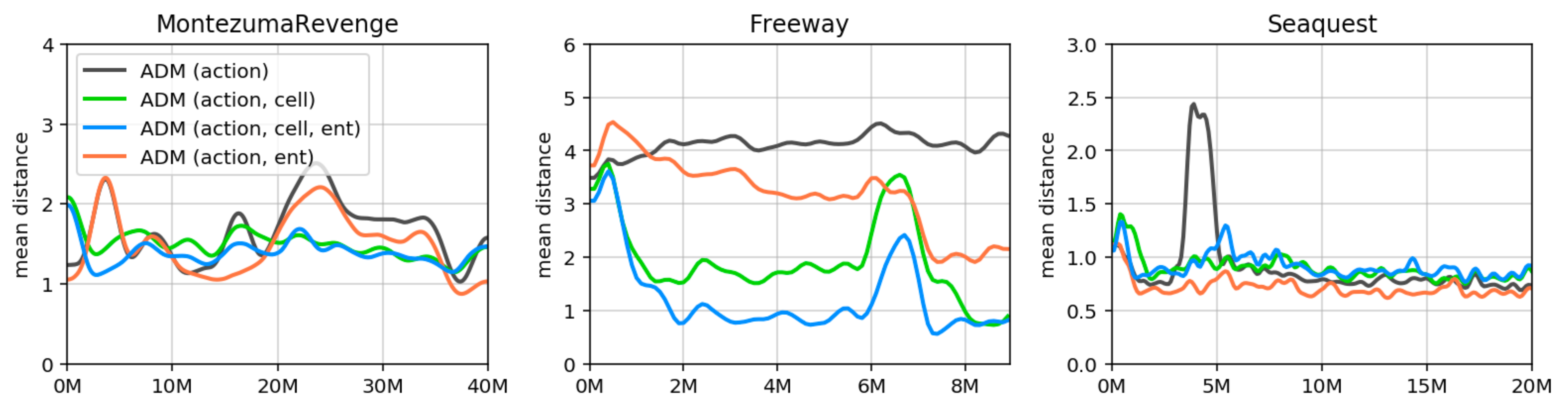}
    \vspace*{-0.2cm}
    \caption{
        Performance of ADM in terms of mean distance
        under different loss combinations in early stages,
        trained using the same online trajectory data.
        Plots were obtained by averaging runs over 5 random seeds.
    }
    \label{fig:a2c_exp_ablation_adm_fixed}
\end{center}
\end{figure*}

Figure~\ref{fig:a2c_exp_ablation_adm_fixed} shows the average distance
between the ground-truth location of the agent and the predicted one by ADM
during the early stages of training.
On \MontezumaRevenge, there is only little difference between the variants
although the full model worked slightly better on average.
On \Freeway, the effect of loss terms is more clear;
in the beginning
the agent tends to behave suboptimally by taking mostly single actions only
(UP out of three action choices --- UP, DOWN, and NO-OP),
hence very low entropy $\mathcal H(\pi(\cdot|s))$,
which can confuse the ADM of telling
which part is actually controllable
as the action classifier would give correct answer regardless of attention.
We can observe additional loss terms help the model
quickly correct the attention %
to localize the controllable object among the uncontrollable clutters
with better stability.

In another ablation study, we compare the end performance of the A2C+\coex{} agent
with the ADM jointly trained under different loss objectives
on these three games (\MontezumaRevenge, \Freeway and \Seaquest).
In our experiments,
the variant with full ADM worked best on \MontezumaRevenge and \Freeway.
The minimal training objective of ADM (\ie, $\mathcal L_\text{action}$)
also solely works reasonably well, but with the combination of other loss terms
we can attain a more stable performance.

\begin{figure*}[th]
\begin{center}
    \includegraphics[width=0.99\linewidth]{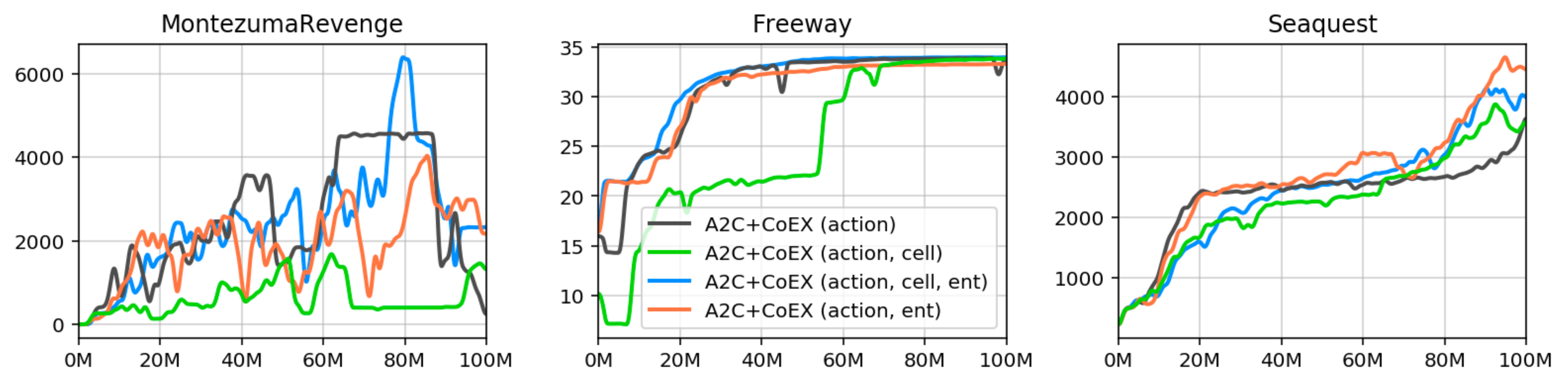}
    \vspace*{-0.2cm}
    \caption{
        Learning curves of A2C+CoEX with ADM trained under different training objectives.
        The curve in solid line shows the mean episode over 40 recent episodes,
        averaged over 3 random seeds.
    }
    \label{fig:a2c_exp_ablation_adm_joint}
\end{center}
\end{figure*}

\clearpage

\section{Ablation Study on the State Representation}
\label{sec:ablation_representation}

We present a result of additional ablation study on the state representation $\psi(s)$ used in count-based exploration.
The following baselines are considered:

\begin{itemize}[leftmargin=7mm]
\setlength{\itemsep}{0pt}\setlength{\parskip}{3pt}
  \item A2C+\coex$(c)$: Uses only the context embedding for exploration,
      \ie, $\psi(s) = (c)$.
  \item A2C+\coex$(c,R)$: Uses only the context embedding and the cumulative reward for exploration
      without contingent region information, \ie, $\psi(s) = (c, R)$.
  \item A2C+\coex$(x,y,c)$: Uses the contingent region information $(x, y)$ as well as the context embedding $c$,
      however without the cumulative reward component, \ie, $\psi(s) = (x, y, c)$.
\end{itemize}

One can also consider another baseline similar to A2C+\coex$(c,R)$
with $\psi(s) = (x, y, c, R)$,
where the location information $(x, y)$ is replaced with random coordinates uniformly sampled from the grid.
It ablates the learned contingent regions.
However, we found that it performs similarly to the presented A2C+CoEX($c, R$) baseline.

\begin{figure*}[t]
\begin{center}
    \includegraphics[width=\linewidth]{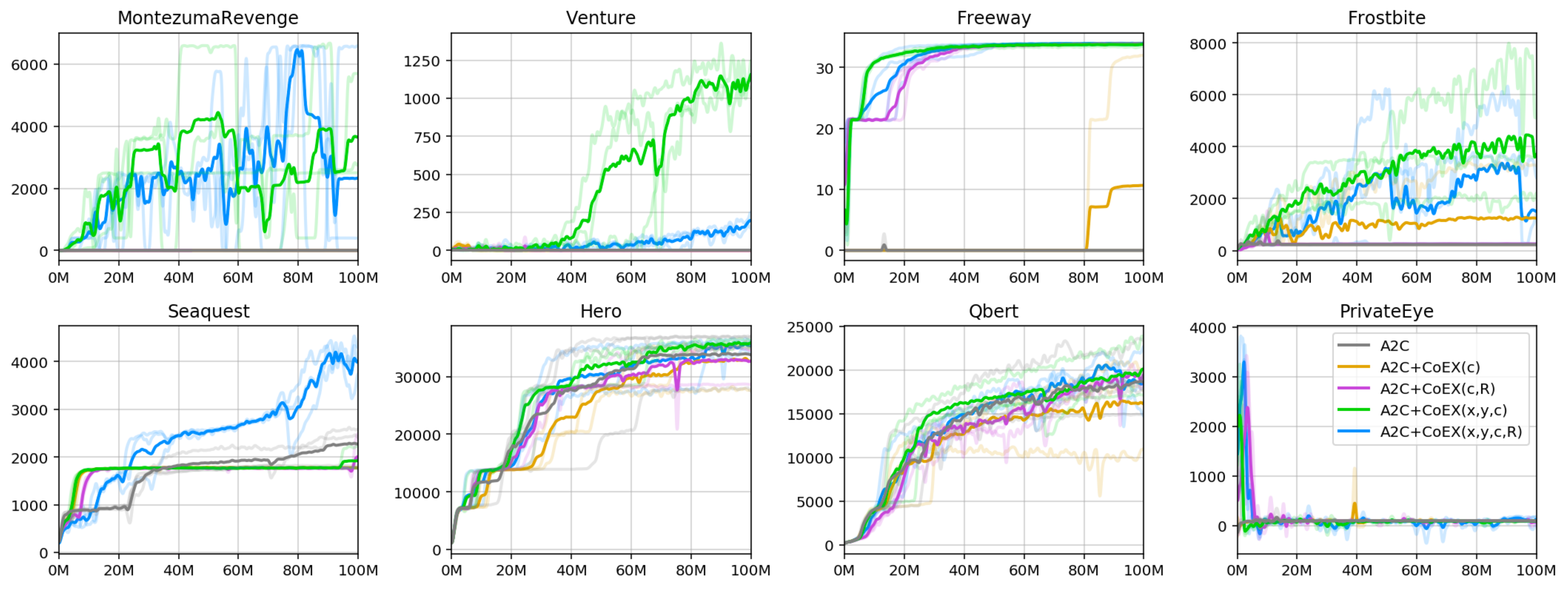}
    \vspace*{-0.5cm}
    \caption{
        Learning curves for the ablation study of state representation.
        The exploration algorithm without the contingent region information (purple)
        performs significantly worse, yielding almost no improvement on hard-exploration games such as
        \MontezumaRevenge, \Venture, and \Frostbite.
        {The mean curve is obtained by averaging over 3 random seeds.}
        See Table \ref{tbl:a2c_additional_performance} for numbers.
    }
    \label{fig:a2c_exp_additional_1}
    \vspace*{-5pt}
\end{center}
\end{figure*}

The experimental results are summarized in Table \ref{tbl:a2c_additional_performance} and
Figure \ref{fig:a2c_exp_additional_1}. %
The variants without contingent regions
(\ie, A2C+\coex($c$) and A2C+\coex($c,R$)
performed significantly worse in most of the games than
A2C+\coex$(x,y,c)$ and
A2C+\coex$(x,y,c,R)$ 
giving little improvement over the A2C baseline.
Most notably, in the games with the hardest exploration such as \MontezumaRevenge and \Venture,
the performance is hardly better than the vanilla A2C or a random policy, achieving a score as low as zero.
The variants with contingent region information worked best and comparable to each other.
We observe that using the cumulative reward (total score) for exploration
gives a slight improvement on some environments.
These results support the effectiveness of the learned contingency-awareness information in count-based exploration.
\renewcommand{\r}{\color{red}}
\begin{table}[t]
    \tabcolsep=0.13cm
    \centering
    \small
    \hspace*{-0.4cm}
    \begin{center}
    \begin{tabular}{@{}l|cccccccc@{}}
        \hline
    \iffalse................................\fi Method & \s Freeway & \s Frostbite & \s Hero  & \s Montezuma & \s PrivateEye & \s Qbert & \s Seaquest & \s Venture \\
        \hline
        A2C                                            &     7.2    & 1099         & 34352    & 12.5         & 574           & 19620    & 2401        & 0          \\
        A2C+\coex{} ($c$)                              &    10.7    & 1313         & 34269    & 14.7         &    2692       & 20942    & 1810        & 94         \\
        A2C+\coex{} ($c,R$)                            & \B 34.0    & 941          & 34046    & 9.2          & \B 5458       & 21587    & 2056        & 77         \\
        A2C+\coex{} ($x,y,c$)                          &    33.7    & \B 5066      &\B 36934  &    6558      &    5377       &    21130 &   1978      & \B 1374    \\
        A2C+\coex{} ($x,y,c,R$)                        & \B 34.0    &    4260      &   36827  & \B 6635      &    5316       & \B 23962 &\B 5169      &    204     \\
        \hline
    \end{tabular}
    \end{center}

    \vspace*{-5pt}
    \caption{
        Summary of the results of the ablation study of the state representation.  %
        We report the maximum mean score (averaged over 40 recent episodes) achieved over 100M environment steps,
        {averaged over 3 random seeds.}
    }
    \label{tbl:a2c_additional_performance}
    \vspace*{-5pt}
\end{table}

%% file: iclr19_coex.bbl
\begin{thebibliography}{45}
\providecommand{\natexlab}[1]{#1}
\providecommand{\url}[1]{\texttt{#1}}
\expandafter\ifx\csname urlstyle\endcsname\relax
  \providecommand{\doi}[1]{doi: #1}\else
  \providecommand{\doi}{doi: \begingroup \urlstyle{rm}\Url}\fi

\bibitem[Achiam \& Sastry(2017)Achiam and Sastry]{Achiam:1703.01732}
Joshua Achiam and Shankar Sastry.
\newblock {Surprise-Based Intrinsic Motivation for Deep Reinforcement
  Learning}.
\newblock \emph{arXiv preprint arXiv:1703.01732}, 2017.

\bibitem[Agrawal et~al.(2016)Agrawal, Nair, Abbeel, Malik, and
  Levine]{Agrawal:2016}
Pulkit Agrawal, Ashvin Nair, Pieter Abbeel, Jitendra Malik, and Sergey Levine.
\newblock {Learning to Poke by Poking: Experiential Learning of Intuitive
  Physics}.
\newblock In \emph{NIPS}, 2016.

\bibitem[Baeyens et~al.(1990)Baeyens, Eelen, and Bergh]{baeyens1990contingency}
Frank Baeyens, Paul Eelen, and Omer van~den Bergh.
\newblock Contingency awareness in evaluative conditioning: A case for unaware
  affective-evaluative learning.
\newblock \emph{Cognition and emotion}, 4\penalty0 (1):\penalty0 3--18, 1990.

\bibitem[Bahdanau et~al.(2015)Bahdanau, Cho, and
  Bengio]{Bahdanau:ICLR2015:Attention}
Dzmitry Bahdanau, Kyunghyun Cho, and Yoshua Bengio.
\newblock {Neural Machine Translation by Jointly Learning to Align and
  Translate}.
\newblock In \emph{ICLR}, 2015.

\bibitem[Banino et~al.(2018)Banino, Barry, Uria, Blundell, Lillicrap, Mirowski,
  Pritzel, Chadwick, Degris, Modayil, Wayne, Soyer, Viola, Zhang, Goroshin,
  Rabinowitz, Pascanu, Beattie, Petersen, Sadik, Gaffney, King, Kavukcuoglu,
  Hassabis, Hadsell, and Kumaran]{Banino:Nature2018:NavigateAI}
Andrea Banino, Caswell Barry, Benigno Uria, Charles Blundell, Timothy
  Lillicrap, Piotr Mirowski, Alexander Pritzel, Martin~J Chadwick, Thomas
  Degris, Joseph Modayil, Greg Wayne, Hubert Soyer, Fabio Viola, Brian Zhang,
  Ross Goroshin, Neil Rabinowitz, Razvan Pascanu, Charlie Beattie, Stig
  Petersen, Amir Sadik, Stephen Gaffney, Helen King, Koray Kavukcuoglu, Demis
  Hassabis, Raia Hadsell, and Dharshan Kumaran.
\newblock {Vector-based navigation using grid-like representations in
  artificial agents}.
\newblock \emph{Nature}, 557\penalty0 (7705):\penalty0 429--433, 2018.

\bibitem[Barto(2013)]{Barto:2013:intrinsic}
Andrew~G Barto.
\newblock {Intrinsic motivation and reinforcement learning}.
\newblock In \emph{Intrinsically motivated learning in natural and artificial
  systems}, pp.\  17--47. Springer, 2013.

\bibitem[Bellemare et~al.(2012)Bellemare, Veness, and
  Bowling]{Bellemare:AAAI2012:Contingency}
Marc~G Bellemare, Joel Veness, and Michael Bowling.
\newblock {Investigating Contingency Awareness Using Atari 2600 Games.}
\newblock In \emph{AAAI}, 2012.

\bibitem[Bellemare et~al.(2013)Bellemare, Naddaf, Veness, and
  Bowling]{Bellemare:JAIR2013:ALE}
Marc~G. Bellemare, Yavar Naddaf, Joel Veness, and Michael Bowling.
\newblock {The Arcade Learning Environment: An Evaluation Platform for General
  Agents}.
\newblock \emph{{Journal of Artificial Intelligence Research 47}}, 2013.

\bibitem[Bellemare et~al.(2016)Bellemare, Srinivasan, Ostrovski, Schaul,
  Saxton, and Munos]{Bellemare:NIPS2016:UnifyingCount}
Marc~G. Bellemare, Sriram Srinivasan, Georg Ostrovski, Tom Schaul, David
  Saxton, and Remi Munos.
\newblock {Unifying Count-Based Exploration and Intrinsic Motivation}.
\newblock In \emph{NIPS}, 2016.

\bibitem[Bengio et~al.(2017)Bengio, Thomas, Pineau, Precup, and
  Bengio]{Bengio:2017:FeatureControl}
Emmanuel Bengio, Valentin Thomas, Joelle Pineau, Doina Precup, and Yoshua
  Bengio.
\newblock {Independently Controllable Features}.
\newblock \emph{arXiv preprint arXiv:1703.07718}, 2017.

\bibitem[Burda et~al.(2018)Burda, Edwards, Pathak, Storkey, Darrell, and
  Efros]{Burda:2018:Curiosity}
Yuri Burda, Harri Edwards, Deepak Pathak, Amos Storkey, Trevor Darrell, and
  Alexei~A. Efros.
\newblock Large-scale study of curiosity-driven learning.
\newblock \emph{arXiv preprint arXiv:1808.04355}, 2018.

\bibitem[Burda et~al.(2019)Burda, Edwards, Storkey, and Klimov]{ICLR2019:RND}
Yuri Burda, Harrison Edwards, Amos Storkey, and Oleg Klimov.
\newblock {Exploration by random network distillation}.
\newblock In \emph{ICLR}, 2019.
\newblock URL \url{https://openreview.net/forum?id=H1lJJnR5Ym}.

\bibitem[Charikar(2002)]{charikar2002}
Moses~S Charikar.
\newblock Similarity estimation techniques from rounding algorithms.
\newblock In \emph{Proceedings of the thiry-fourth annual ACM symposium on
  Theory of computing}, pp.\  380--388. ACM, 2002.

\bibitem[Dhariwal et~al.(2017)Dhariwal, Hesse, Klimov, Nichol, Plappert,
  Radford, Schulman, Sidor, and Wu]{Dhariwal:2017:baselines}
Prafulla Dhariwal, Christopher Hesse, Oleg Klimov, Alex Nichol, Matthias
  Plappert, Alec Radford, John Schulman, Szymon Sidor, and Yuhuai Wu.
\newblock {OpenAI Baselines}.
\newblock \url{https://github.com/openai/baselines}, 2017.

\bibitem[Dilokthanakul et~al.(2017)Dilokthanakul, Kaplanis, Pawlowski, and
  Shanahan]{Dilokthanakul:2017:IntrinsicMotivation}
Nat Dilokthanakul, Christos Kaplanis, Nick Pawlowski, and Murray Shanahan.
\newblock {Feature Control as Intrinsic Motivation for Hierarchical
  Reinforcement Learning}.
\newblock \emph{arXiv preprint arXiv:1705.06769}, 2017.

\bibitem[Durrant-Whyte \& Bailey(2006)Durrant-Whyte and
  Bailey]{Durrant:2006:SLAM}
Hugh Durrant-Whyte and Tim Bailey.
\newblock {Simultaneous Localization and Mapping: Part I}.
\newblock \emph{IEEE robotics \& automation magazine}, 13\penalty0
  (2):\penalty0 99--110, 2006.

\bibitem[Eysenbach et~al.(2018)Eysenbach, Gupta, Ibarz, and
  Levine]{Eysenbach:2018uc}
Benjamin Eysenbach, Abhishek Gupta, Julian Ibarz, and Sergey Levine.
\newblock {Diversity is All You Need: Learning Skills without a Reward
  Function}.
\newblock 2018.

\bibitem[Houthooft et~al.(2016)Houthooft, Chen, Duan, Schulman, De~Turck, and
  Abbeel]{Houthooft:NIPS2016:VIME}
Rein Houthooft, Xi~Chen, Yan Duan, John Schulman, Filip De~Turck, and Pieter
  Abbeel.
\newblock {VIME: Variational Information Maximizing Exploration}.
\newblock In \emph{NIPS}, 2016.

\bibitem[Jaderberg et~al.(2017)Jaderberg, Mnih, Czarnecki, Schaul, Leibo,
  Silver, and Kavukcuoglu]{Jaderberg:ICLR2017:RLAux}
Max Jaderberg, Volodymyr Mnih, Wojciech~Marian Czarnecki, Tom Schaul, Joel~Z
  Leibo, David Silver, and Koray Kavukcuoglu.
\newblock {Reinforcement Learning with Unsupervised Auxiliary Tasks}.
\newblock In \emph{ICLR}, 2017.

\bibitem[Kulis \& Jordan(2012)Kulis and Jordan]{Kulis:ICML2012:BNPCluster}
Brian Kulis and Michael I~. Jordan.
\newblock {Revisiting k-means: New Algorithms via Bayesian Nonparametrics}.
\newblock In \emph{ICML}, 2012.

\bibitem[Machado et~al.(2017)Machado, Bellemare, Talvitie, Veness, Hausknecht,
  and Bowling]{machado2017:revisiting}
Marlos~C Machado, Marc~G Bellemare, Erik Talvitie, Joel Veness, Matthew
  Hausknecht, and Michael Bowling.
\newblock Revisiting the arcade learning environment: Evaluation protocols and
  open problems for general agents.
\newblock \emph{Journal of Artificial Intelligence Research}, 61:\penalty0
  523--562, 2017.

\bibitem[Martin et~al.(2017)Martin, Sasikumar, Everitt, and
  Hutter]{Martin:1706.08090}
Jarryd Martin, Suraj~Narayanan Sasikumar, Tom Everitt, and Marcus Hutter.
\newblock {Count-Based Exploration in Feature Space for Reinforcement
  Learning}.
\newblock In \emph{IJCAI}, 2017.

\bibitem[Martins \& Astudillo(2016)Martins and
  Astudillo]{Martins:2016:Sparsemax}
Andr{\'e} F~T Martins and Ram{\'o}n~Fernandez Astudillo.
\newblock {From Softmax to Sparsemax: A Sparse Model of Attention and
  Multi-Label Classification}.
\newblock In \emph{ICML}, 2016.

\bibitem[Mirowski et~al.(2017)Mirowski, Pascanu, Viola, Soyer, Ballard, Banino,
  Denil, Goroshin, Sifre, Kavukcuoglu, Kumaran, and
  Hadsell]{Mirowski:ICLR2017:Navigate}
Piotr Mirowski, Razvan Pascanu, Fabio Viola, Hubert Soyer, Andrew~J Ballard,
  Andrea Banino, Misha Denil, Ross Goroshin, Laurent Sifre, Koray Kavukcuoglu,
  Dharshan Kumaran, and Raia Hadsell.
\newblock {Learning to Navigate in Complex Environments}.
\newblock In \emph{ICLR}, 2017.

\bibitem[Mirowski et~al.(2018)Mirowski, Grimes, Malinowski, Hermann, Anderson,
  Teplyashin, Simonyan, Kavukcuoglu, Zisserman, and
  Hadsell]{Mirowski:NIPS2018:StreetLearn}
Piotr Mirowski, Matthew~Koichi Grimes, Mateusz Malinowski, Karl~Moritz Hermann,
  Keith Anderson, Denis Teplyashin, Karen Simonyan, Koray Kavukcuoglu, Andrew
  Zisserman, and Raia Hadsell.
\newblock {Learning to Navigate in Cities Without a Map}.
\newblock In \emph{NIPS}, 2018.

\bibitem[Mnih et~al.(2015)Mnih, Kavukcuoglu, Silver, Rusu, Veness, Bellemare,
  Graves, Riedmiller, Fidjeland, Ostrovski, Petersen, Beattie, Sadik,
  Antonoglou, King, Kumaran, Wierstra, Legg, and Hassabis]{Mnih:Nature2015:DQN}
Volodymyr Mnih, Koray Kavukcuoglu, David Silver, Andrei~A. Rusu, Joel Veness,
  Marc~G. Bellemare, Alex Graves, Martin Riedmiller, Andreas~K. Fidjeland,
  Georg Ostrovski, Stig Petersen, Charles Beattie, Amir Sadik, Ioannis
  Antonoglou, Helen King, Dharshan Kumaran, Daan Wierstra, Shane Legg, and
  Demis Hassabis.
\newblock {Human-level control through deep reinforcement learning}.
\newblock \emph{Nature}, 2015.

\bibitem[Mnih et~al.(2016)Mnih, Badia, Mirza, Graves, Lillicrap, Harley,
  Silver, and Kavukcuoglu]{Mnih:ICML2016:A3C}
Volodymyr Mnih, Adri{\`a}~Puigdom{\`e}nech Badia, Mehdi Mirza, Alex Graves,
  Timothy~P Lillicrap, Tim Harley, David Silver, and Koray Kavukcuoglu.
\newblock {Asynchronous Methods for Deep Reinforcement Learning}.
\newblock In \emph{ICML}, 2016.

\bibitem[Moser et~al.(2015)Moser, Rowland, and I~Moser]{Moser:2015:GridCell}
May-Britt Moser, David Rowland, and Edvard I~Moser.
\newblock {Place Cells, Grid Cells, and Memory}.
\newblock \emph{Cold Spring Harbor perspectives in medicine}, 5, 2015.

\bibitem[Oh et~al.(2015)Oh, Guo, Lee, Lewis, and
  Singh]{Oh:NIPS2015:ActionVideoPred}
Junhyuk Oh, Xiaoxiao Guo, Honglak Lee, Richard~L Lewis, and Satinder~P Singh.
\newblock {Action-Conditional Video Prediction using Deep Networks in Atari
  Games.}
\newblock In \emph{NIPS}, 2015.

\bibitem[Osband et~al.(2016)Osband, Blundell, Pritzel, and
  Van~Roy]{Osband:NIPS2016:BootstrappedDQN}
Ian Osband, Charles Blundell, Alexander Pritzel, and Benjamin Van~Roy.
\newblock {Deep Exploration via Bootstrapped DQN}.
\newblock In \emph{NIPS}, 2016.

\bibitem[Ostrovski et~al.(2017)Ostrovski, Bellemare, van~den Oord, and
  Munos]{Ostrovski:ICML2017:ExplorationDensity}
Georg Ostrovski, Marc~G. Bellemare, Aaron van~den Oord, and Remi Munos.
\newblock {Count-Based Exploration with Neural Density Models}.
\newblock In \emph{ICML}, 2017.

\bibitem[Oudeyer \& Kaplan(2009)Oudeyer and Kaplan]{Oudeyer:2009}
Pierre-Yves Oudeyer and Frederic Kaplan.
\newblock {What is intrinsic motivation? A typology of computational
  approaches}.
\newblock \emph{Frontiers in Neurorobotics}, 2009.

\bibitem[Pathak et~al.(2017)Pathak, Agrawal, Efros, and
  Darrell]{Pathak:ICML2017:Curiosity}
Deepak Pathak, Pulkit Agrawal, Alexei~A Efros, and Trevor Darrell.
\newblock {Curiosity-driven Exploration by Self-supervised Prediction}.
\newblock In \emph{ICML}, 2017.

\bibitem[Plappert et~al.(2018)Plappert, Houthooft, Dhariwal, Sidor, Chen, Chen,
  Asfour, Abbeel, and Andrychowicz]{Plappert:ICLR2018:ParamExp}
Matthias Plappert, Rein Houthooft, Prafulla Dhariwal, Szymon Sidor, Richard~Y
  Chen, Xi~Chen, Tamim Asfour, Pieter Abbeel, and Marcin Andrychowicz.
\newblock {Parameter Space Noise for Exploration}.
\newblock In \emph{ICLR}, 2018.

\bibitem[Rand(1971)]{Rand:ARI}
William~M. Rand.
\newblock {Objective criteria for the evaluation of clustering methods}.
\newblock \emph{Journal of the American Statistical Association}, 66\penalty0
  (336):\penalty0 846--850, 1971.

\bibitem[Sawada(2018)]{Sawada:2018:Controllable}
Yoshihide Sawada.
\newblock {Disentangling Controllable and Uncontrollable Factors of Variation
  by Interacting with the World}.
\newblock \emph{arXiv preprint arXiv:1804.06955}, 2018.

\bibitem[Schmidhuber(1991)]{Schmidhuber:1991}
Jürgen Schmidhuber.
\newblock Adaptive confidence and adaptive curiosity.
\newblock 1991.

\bibitem[Schulman et~al.(2017)Schulman, Wolski, Dhariwal, Radford, and
  Klimov]{schulman2017:proximal}
John Schulman, Filip Wolski, Prafulla Dhariwal, Alec Radford, and Oleg Klimov.
\newblock Proximal policy optimization algorithms.
\newblock \emph{arXiv preprint arXiv:1707.06347}, 2017.

\bibitem[Shelhamer et~al.(2017)Shelhamer, Mahmoudieh, Argus, and
  Darrell]{Shelhamer:2017}
Evan Shelhamer, Parsa Mahmoudieh, Max Argus, and Trevor Darrell.
\newblock {Loss is its own Reward: Self-Supervision for Reinforcement
  Learning}.
\newblock \emph{arXiv preprint arXiv:1612.07307}, 2017.

\bibitem[Singh et~al.(2004)Singh, Chentanez, and
  Barto]{Chentanez:NIPS2004:Intrinsic}
Satinder Singh, Nuttapong Chentanez, and Andrew~G. Barto.
\newblock {Intrinsically Motivated Reinforcement Learning}.
\newblock In \emph{NIPS}, 2004.

\bibitem[Strehl \& Littman(2008)Strehl and Littman]{Strehl:2008}
Alexander~L. Strehl and Michael~L. Littman.
\newblock An analysis of model-based interval estimation for markov decision
  processes.
\newblock \emph{Journal of Computer and System Sciences}, 74\penalty0 (8),
  2008.

\bibitem[Tang et~al.(2017)Tang, Houthooft, Foote, Stooke, Chen, Duan, Schulman,
  Turck, and Abbeel]{Tang:NIPS2017:SimHash}
Haoran Tang, Rein Houthooft, Davis Foote, Adam Stooke, Xi~Chen, Yan Duan, John
  Schulman, Filip~De Turck, and Pieter Abbeel.
\newblock {\#Exploration: A Study of Count-Based Exploration for Deep
  Reinforcement Learning}.
\newblock In \emph{NIPS}, 2017.

\bibitem[Watson(1966)]{watson1966:contingency}
John~S Watson.
\newblock {The development and generalization of "contingency awareness" in
  early infancy: Some hypotheses}.
\newblock \emph{Merrill-Palmer Quarterly of Behavior and Development},
  12\penalty0 (2):\penalty0 123--135, 1966.

\bibitem[Xu et~al.(2015)Xu, Ba, Kiros, Cho, Courville, Salakhutdinov, Zemel,
  and Bengio]{Xu:ICML2015:ShowAttendTell}
Kelvin Xu, Jimmy Ba, Ryan Kiros, Kyunghyun Cho, Aaron~C. Courville, Ruslan
  Salakhutdinov, Richard~S Zemel, and Yoshua Bengio.
\newblock {Show, Attend and Tell: Neural Image Caption Generation with Visual
  Attention.}
\newblock In \emph{ICML}, 2015.

\bibitem[Zheng et~al.(2018)Zheng, Oh, and Singh]{Zheng:NIPS2018:LIRPG}
Zeyu Zheng, Junhyuk Oh, and Satinder Singh.
\newblock {On Learning Intrinsic Rewards for Policy Gradient Methods}.
\newblock In \emph{NIPS}, 2018.

\end{thebibliography}
